\documentclass[11pt,a4paper]{article}
\usepackage[final]{emnlp2021}
\usepackage{times}
\usepackage{latexsym}

\usepackage{graphicx}
\usepackage{tikz}
\usepackage{subfig}
\usepackage{mdframed}
\usepackage{tabularx}
\usepackage{amssymb}
\usepackage{amsmath}
\usepackage{verbatim}
\usepackage{hyperref}
\usepackage{nert}

\usepackage{microtype}

\definecolor{brickred}{HTML}{8f1402}
\definecolor{darkgreen}{HTML}{066900}

\title{Modeling Disclosive Transparency in NLP Application Descriptions}

\author{Michael Saxon, Sharon Levy, Xinyi Wang, Alon Albalak, William Yang Wang \\
  University of California, Santa Barbara \\
  \texttt{\small\{\emldisplay{saxon@ucsb.edu}{saxon}, \emldisplay{xinyi\_wang@ucsb.edu}{xinyi\_wang}, \emldisplay{alon\_albalak@ucsb.edu}{alon\_albalak}\}@ucsb.edu, \{\emldisplay{sharonlevy@ucsb.edu}{sharonlevy}, \emldisplay{william@cs.ucsb.edu}{william}\}@cs.ucsb.edu}
  }

\date{}

\begin{document}
\maketitle
\begin{abstract}
Broader disclosive transparency---truth and clarity in communication regarding the function of AI systems---is widely considered desirable. Unfortunately, it is a nebulous concept, difficult to both define and quantify. This is problematic, as previous work has demonstrated possible trade-offs and negative consequences to disclosive transparency, such as a confusion effect, where ``too much information'' clouds a reader's understanding of what a system description means. Disclosive transparency's subjective nature has rendered deep study into these problems and their remedies difficult. To improve this state of affairs, We introduce neural language model-based probabilistic metrics to directly model disclosive transparency, and demonstrate that they correlate with user and expert opinions of system transparency, making them a valid objective proxy. Finally, we demonstrate the use of these metrics in a pilot study quantifying the relationships between transparency, confusion, and user perceptions in a corpus of real NLP system descriptions. 

\end{abstract}

\section{Introduction}\label{sec:intro}

Among the draft \textit{Ethics Guidelines for Trustworthy AI} released by the European Union in 2019 were calls for greater transparency around deployed systems using artificial intelligence, advising that an ``AI system’s capabilities and limitations should be communicated to practitioners or end-users in a manner appropriate to the use case at hand.'' \cite{ec2019ethics} 
This is a high-profile example of the vagueness endemic to guidance on ethical disclosure and AI---\textit{what constitutes an appropriate manner of communication}?  

There is a growing awareness of the importance of communicating AI system function and performance clearly and understandably. This is both a matter of public interest, for mitigating harms caused by, for example, racial biases in the performance of human classifiers \cite{raji2019actionable}, but also in the interest of system providers, as users who feel like they do not understand how machine learning models work and what kind of information they rely on tend to be more resistant to using them \cite{poursabzi2018manipulating}. 

However, while fairness \cite{pmlr-v81-dwork18a, harrison_empirical_2020} and privacy as general principles \cite{ji2014differential, papernot2016towards} have been well-studied in the context of AI, transparency does not receive as much attention. The term \textit{transparency} is overloaded \cite{lipton2019research}, with many different definitions in the literature \cite{felzmann_robots_2019}. It has a broad meaning in the public consciousness, from which some studies adopt a ``you'll know it when you see it'' definition \cite{doshi2017towards}. 
As opposed to the notions of \textit{transparency as explainability} \cite{veale_2018_fairness} and \textit{as invisibility} \cite{hamilton_path_2014}, \textit{transparency as disclosure} \cite{suzor_what_2019}, the extent to which the producers of an AI system or service provide detailed messaging about a system to stakeholders such as buyers, users, or the general public. This third sense, \textit{\textbf{disclosive transparency}}, is particularly subjective.

\citet{pieters2011explanation} identifies a confusion dynamic where providing ``too much information'' about a decision system worsens user understanding and trust. 
This dynamic could have serious ramifications---if disclosive transparency really decreases user trust, stakeholders are further incentivized to not disclose. Unfortunately, studies on this relationship are small-scale and subjective. This is why measurable, repeatable, and objective measures of disclosive transparency are sorely needed. In this work, we \textbf{decompose disclosive transparency} into two components: \textit{replicability} and \textit{style-appropriateness}, \textbf{develop three objective neural language model measures} of them, and apply them in a \textbf{pilot study of real systems}\footnote{All code, data, and annotations are available online at \texttt{ \href{https://github.com/michaelsaxon/disclosive-transparency}{github.com/michaelsaxon/disclosive-transparency}}.}.


\section{Decomposing Disclosive Transparency}

Our goal is to develop an objective measure of disclosive transparency that is repeatable, explainable, intuitive, and well-correlated with subjective opinions. However, at face value there is not an obvious way to assign a numerical value to the ``level of transparency'' in a description. To resolve some vagueness, we decompose disclosive transparency into two components: the \textbf{replicability}, or degree to which the requisite content to reproduce the system being described is present in a description, and the \textbf{style-appropriateness}, or the degree to which it is written in a manner understandable to a member of the general public. These components are more specific than ``transparency'' alone, but still quite subjective.

A key insight underpins the possibility of developing objective measures of replicability and style-appropriateness: \textbf{disclosure is a communication task}. In describing a system, an explainer, typically an authority  providing an AI service to the public, attempts to encode information about the design and function of their system in a \textit{summary}. We find the communication-based definition of meaning provided in \citet{bender-koller-2020-climbing} useful for motivation.  In their framework, the \textit{meaning}, $m$ of an utterance is a pair $(e, i)$ of the surface form $e$ (such as text or speech audio) and \textit{communicative intent}, $i$ which is external to language. In the case of disclosive transparency, this $i$ is the particulars of the system being described.  Any assessment of the disclosive transparency of a description $e$ is fundamentally an assessment of its underlying $i$---the degree to which $i$ contains the information necessary to reconstruct the system being described. 

Furthermore, \textbf{some short descriptions come paired with longer ones}. Anyone who has clicked ``I Agree'' without reading a terms of service, but has taken the time to look over a short, to-the-point statement on how data is used, knows that the lengthy legalese-laden descriptions that truly define systems we interact with can be sufficiently summarized succinctly. However, the succinct summarization is intended to carry the detailed meaning of the actual contract a user is agreeing to. In other words, for the short description $e'$ of an agreement, there exists a `source' document $e$ that is much more detailed, which share a common $i$. An analogue in system descriptions is what enables our metrics.

\section{Objective Transparency Metrics}




\begin{figure}[t]
    \centering
    \includegraphics[width=\linewidth]{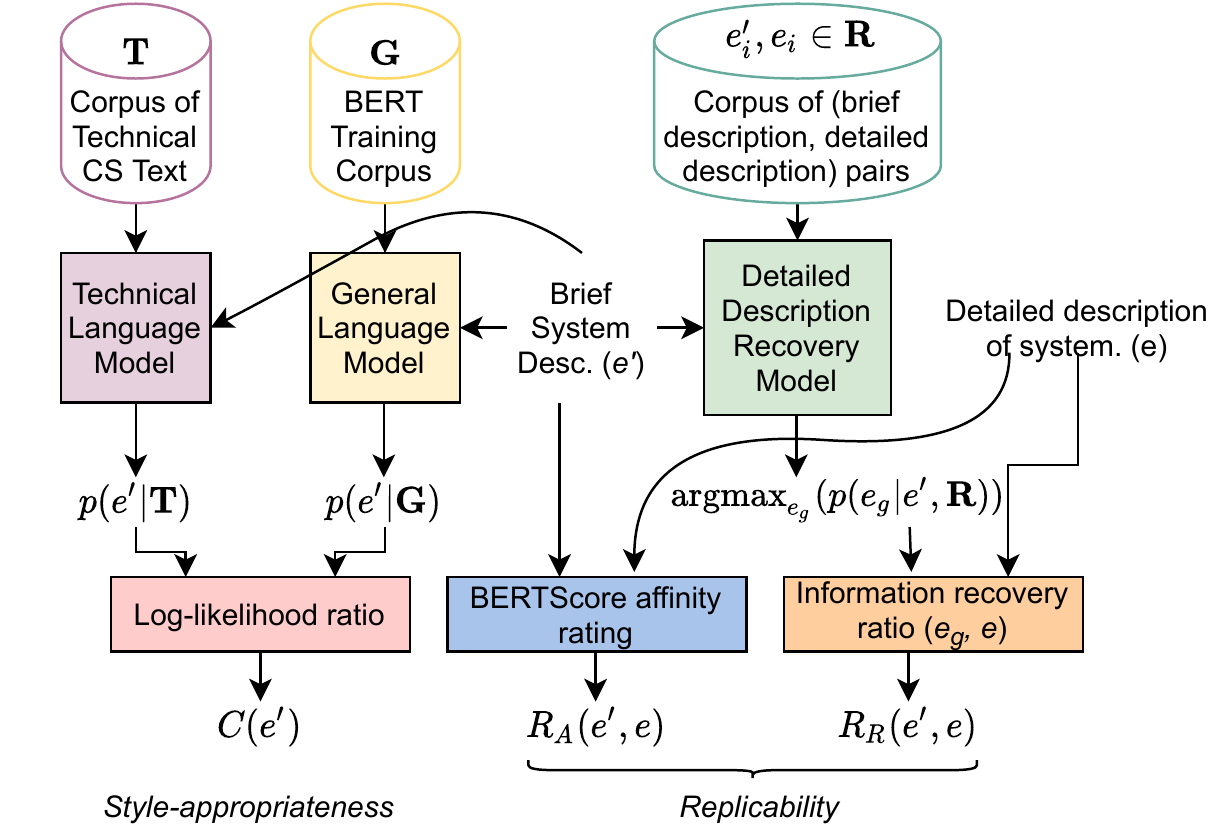} 
	\caption{A diagram of our proposed method for extracting the \textit{style-appropriateness} and \textit{replicability} objective features to characterize disclosive transparency.
	}
	\label{fig:method}
\end{figure}


We now move to developing usable metrics that quantify both the replicability and style-appropriateness of real system descriptions. The system demonstration tracks at academic AI conferences represent a good source of real-world system descriptions. Ideally, the purpose of a system demonstration paper at an academic AI venue is to provide a sufficiently detailed system description such that a peer could (given sufficient resources and data) replicate it. Each of these lengthy documents, $e_i$, is accompanied by an abstract, $e'_i$ which briefly describes the system. Although these abstracts are not intended for consumption by the general public, they do succinctly describe implemented systems, some of which are intended to (at least hypothetically) be used by the public. Combined with the fact that these paper/abstract pairs are freely available online, these are an appealing subject of study for modeling disclosive transparency. Details on our pilot study using demo track NLP papers is provided in \autoref{sec:study}.


\autoref{fig:method} provides a high-level depiction of our objective disclosive transparency metrics, the \textit{style-appropriateness} `clarity' metric $C(e)$, and the \textit{replicability metrics} of `sentence affinity' $R_A(e', e)$ and simulated `information recovery ratio' $R_R(e',e)$. We assess these features using language model scores derived from GPT-2 \cite{radford2019language} and BERT \cite{devlin2018bert}. 

For fine-tuning data on academic language in AI and CS, we produce a training dataset by crawling a random sample of 100k \LaTeX\space files from the arXiv preprint repository with topic label cs.* from 2007-2020. Additionally, we further collect all 30k crawlable \LaTeX\space files from the cs.CL label from 2017-2020, to adequately sample recent research in natural language processing and computational linguistics. From these raw \texttt{.tex} files we produce task-specific plaintext training corpora as described in Appendix C. We then fine-tune the GPT-2 regular pretrained model from Huggingface \cite{wolf2020transformers} on the corpora as explained below. We manually exclude all papers used in our downstream case study from pretraining.

\begin{figure*}[t!]
	\centering
     \includegraphics[width=\linewidth]{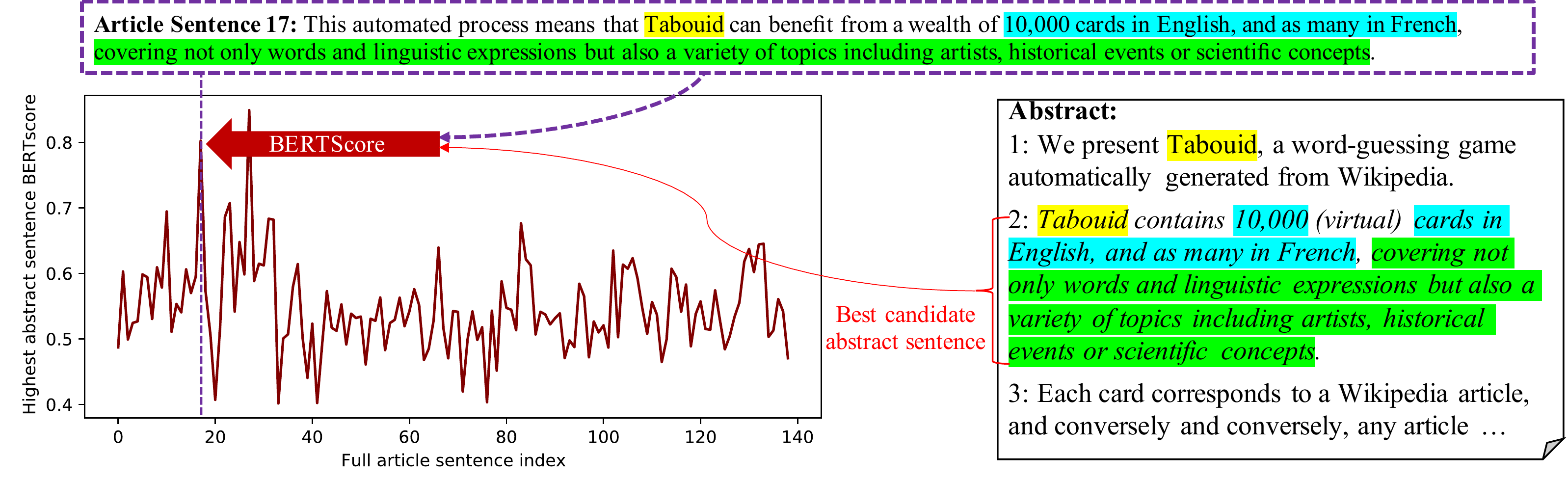} 
 	\caption{A demonstration of how our \textit{sentence affinity} curve simulates the replicability of the abstract based on the degree to which it ``covers'' the content of the full article. For each line in the source document, the maximum pairwise BERTscore between it and the sentences in the short system description is measured. The affinity-replicability score of the article is the paper length-normalized area under this curve.}
	\label{fig:abs}
\end{figure*}

\subsection{Modeling Replicability}
When discussing system descriptions, replicability typically refers to the ability of a third party to implement a functionally equivalent replacement system from a description. Rather than automate replication itself, we use the process of \textit{recovering the full text of the system description from the abstract} as a proxy for replicability.

As we established previously, a reasonable assumption for the communicative goal $i$ of an author of an academic paper $e$ describing a system is to transmit the information necessary to reconstruct it. Obviously, it is not generally possible to perfectly recover $i$ from $e'$ or academic papers beyond the abstract would be generally superfluous.

The problem of generalized inversion of communicative intent $i$ from form $e$ is yet unsolved \cite{yogatama2019learning}. So, we treat recovery of $e$ from a short summary $e'$ as a proxy for recovering $i$, reconstructing the system. In other words, \textbf{the replicability of an abstract $e'$ can be modeled as the amount of information contained in the full article text $e$ than can be recovered from $e'$}. We propose two metrics to simulate this process using the document $e$ and abstract $e'$; \textit{trigram information recovery} and \textit{sentence affinity}. 




\subsubsection{Trigram recovery}

We directly attempt the recovery of $e$ from $e'$ using a generative language model (in this case, GPT-2 \cite{radford2019language}) fine-tuned on a \textit{full-text recovery task}, where for each abstract the replicability score is the rate of trigram information in $e$ recovered by the model-generated text. 

The model generates predicted sentences $e_g$ from the full paper conditioned on an abstract. Given promising results on quantifying semantic complexity using simple measures such as n-gram entropy \cite{mckenna2020semantic}, we use a trigram self-information content metric to measure the amount of information content in $e'$ that is recovered in $e_g$.

Using the training dataset global trigram distribution, we take the ratio of the trigram self-information of all trigrams present in $e_g$ that were recovered from $e$ against the total trigram self-information of $e$. This gives us a ratio of recovered form information $R$ as defined in \autoref{eqn:R}: 

\begin{equation}\label{eqn:R}
    R_R(e, e_g) = \frac{\sum_{t\in e_{g} \cap e} \log(p(t))}{\sum_{t\in e} \log(p(t))}
\end{equation}

\subsubsection{Sentence affinity}

Using the aforementioned trigram recovery metric to model replicability has risks. The distribution of source papers could be too specific, and not contain requisite information, or the generated output might be too noisy to meaningfully simulate the process of inverting system from abstract. Thus, we present an alternative approach here.

\citet{zhang2019bertscore} presented a method for evaluating the similarity of sentence pairs called BERTscore, which is computed by averaging a greedy match of the cosine similarities of the BERT \cite{devlin2018bert} token embeddings $\textbf{b}(t)$ for each token in a pair of sentences:

\begin{equation}
    P_\text{BERT}(s_1,s_2) = \frac{1}{|s_1|}\sum_{t_i\in s_1}(\max_{t_j \in s_2}(\textbf{b}(t_i)^T\textbf{b}(t_j))
\end{equation}

We propose extending this to match a sentence over a set of sentences in a document to produce a document-level \textit{affinity score},

\begin{equation}
    R_A(e_1, e_2) = \frac{1}{|e_1|}\sum_{s_1\in e_1}\max_{s_2 \in e_2}(P_\text{BERT}(s_1, s_2))
\end{equation}

This metric has advantages over trigram recovery in that it doesn't rely on training a generative language model, and is more readily interpretable, as \autoref{fig:abs} demonstrates.

\subsection{Modeling Style-appropriateness}

Orthogonal to the question of whether an abstract contains the necessary information to perform a replication is the question of \textbf{who would be capable of performing the replication}. This question is important for assessing the appropriateness of a description to layperson audiences. We simulate this with a pair of language models tuned to two styles of writing---one general interest, one scientific---to make a perceptual model of ``academic style,'' as a layperson would probably require many more textbook lookups to understand an academic paper or detailed terms of service than they would reading a Wikipedia article. This approach is analogous to a popular the use of perceptual likelihood ratios in speech processing, which have been shown to correlate well with subjective perceptual opinions \cite{saxon2020robust}.

For each passage we extract a likelihood ratio between two language models, one a GPT-2 model fine-tuned on the 100k file arXiv corpus, the other the vanilla GPT-2 model that is pretrained on a large, diverse corpus of online English text. Given the system demo abstract $e$, we compute the style-appropriateness $C$ as a log-likelihood ratio that it belongs to this academia-specific distribution $A$ or a general public distribution $V$ as follows:
\begin{equation}\label{eqn:C}
    C(e) = - \sum_{j=1}^{|\{t|t\in e\}|} \log\left(\frac{p(t_j|A, t_{j-1}...t_1)}{p(t_j|V, t_{j-1}...t_1)}\right)  \\
\end{equation}



\section{Pilot Studies}
\label{sec:study}
To analyze how well our objective measures track with subjective notions of transparency, and to demonstrate how they might be used to investigate how transparency affects prospective user opinions in the real world, we perform two pilot studies utilizing our metrics and a corpus of real-world AI system descriptions.

\subsection{*ACL System Demo Corpus}

We extract system demonstration abstracts from EMNLP 2017--2020, ACL 2018--2020, and NAACL 2018 and 2019, retrieving a corpus of 268 abstracts describing a variety of demonstrations, including systems intended for use by the general public (e.g., translation systems, newsreaders) as well as demonstrations that are of interest more narrowly to the NLP community, software developers, or academics at large (e.g., toolkits, packages, or benchmarks). As we are interested in system descriptions for non-experts, \textbf{we restrict our analysis to abstracts which describe systems intended for use by laypeople}. This set contains 55 abstracts, describing diverse systems from automated language learning games, to news aggregators, to specialized search engines for medical topics. 

We first collect expert opinions on how the abstracts conform to the aforementioned dimensions of transparency to analyze the quality of our automated  metrics. Then, we collect salient layperson opinions of trust, understanding, and fairness to demonstrate a pilot study of how transparency drives user attitudes. 

\subsection{Connecting Objective with Subjective}\label{sec:dims}

Our first pilot study seeks to determine the extent to which our objective measures faithfully model experts' subjective notions of transparency. To do this we must first pose a set of  precise questions for subjectively assessing  disclosive transparency. We identify three largely disjoint \textit{dimensions of disclosive transparency} that each follow directly from a key question an implementer might ask when initially trying to understand a description:

\paragraph{Task Transp.:} What task does this system solve?

\paragraph{Function Transp.:} What components does this system contain, and how do they work?

\paragraph{Data Transp.:} What are the inputs and outputs of this system? What kind of data collection and storage is required to train and operate it?

\paragraph{}
Each of these questions concerns some kind of information contained in $i$. They can be posed as survey questions by appending ``to what extent does the description explain...'' to the start.

We consider \textbf{task} transparency interesting because members of the general public are often learning of new system application areas in which emerging technologies such as NLP can be applied, and without communicating the parameters of the task a system is solving clearly, users cannot understand it.  \textbf{Function} transparency is perhaps the most natural dimension, as discussing the ``how'' of a system is fundamental to explaining how it works. Finally, we consider \textbf{data} transparency because many public discussions of AI ethics center on the use and misuse of data, and it is a central focus of regulatory discussions of AI.

\subsubsection{Collecting the Subjective Ratings}

Four NLP Ph.D. students provided five-point Likert opinion scores of the task, function, and data transparency levels for each abstract in the *ACL Corpus. To ensure consistency across the raters we analyzed average abstract-wise variance, as well as the average pairwise inter-rater Pearson correlation coefficient (PCC) and \textit{p}-value for each of the three transparency categories. \autoref{tab:irr} demonstrates the high inter-rater reliability of the scores, with each average variance $<0.55$ on the 5-point scale.


\begin{table}[t]
\centering
\small
\begin{tabular}{lccc}
\hline
\textbf{Transparency} & \textbf{Variance} & \textbf{PCC} & \textbf{\textit{p}}\\
\hline
Task & 0.428 & 0.508 &	0.001 \\
Function & 0.506 &  0.319 &	0.085\\
Data & 0.488 & 0.368 &	0.016\\\hline
\end{tabular}
\caption{Inter-rater reliability assessed using variance, average pairwise Pearson's correlation coefficient (PCC) and \textit{p}-value for subjective transparency scores.}\label{tab:irr}
\end{table}


Each abstract in the dataset is assigned subjective task, function, and data transparency scores by averaging the four opinion ratings.

\subsection{Pilot Study of User Attitudes}

In this section we briefly outline how we assess user opinions regarding the system descriptions.

Starting from the aforementioned dimensions of transparency---task, function, and data---we find three sets of user concerns orthogonal to these dimensions. These are ``understanding,'' ``fairness,'' and ``trust.'' Together these form ``user response variables'' such as task understanding, function fairness, or data trust, which can be posed as questions about what the user believes or understands. 

To measure how user attitudes and confusion correlate with description transparency, we simulate a user survey study using Amazon Mechanical Turk (AMT). Each abstract was shown to 10 crowd workers selected from a set of majority English-speaking locales (US, CA, AU, NZ, IE, UK), who were instructed to read the abstract and answer two sets of multiple-choice questions. The first set, \textit{opinion prompts}, consists of five-point Likert scale subjective attitude questions; the second set, of \textit{retention questions}, reveals how well the users can recall phrases from the abstract they just read. See Appendix B for survey methodology details.

\section{Results}



\begin{figure*}[t]
\begin{minipage}{.33\linewidth}
\centering
\label{subfig:d_f}
\subfloat[$\textrm{PCC}=0.309$]{\includegraphics[width=1\linewidth]{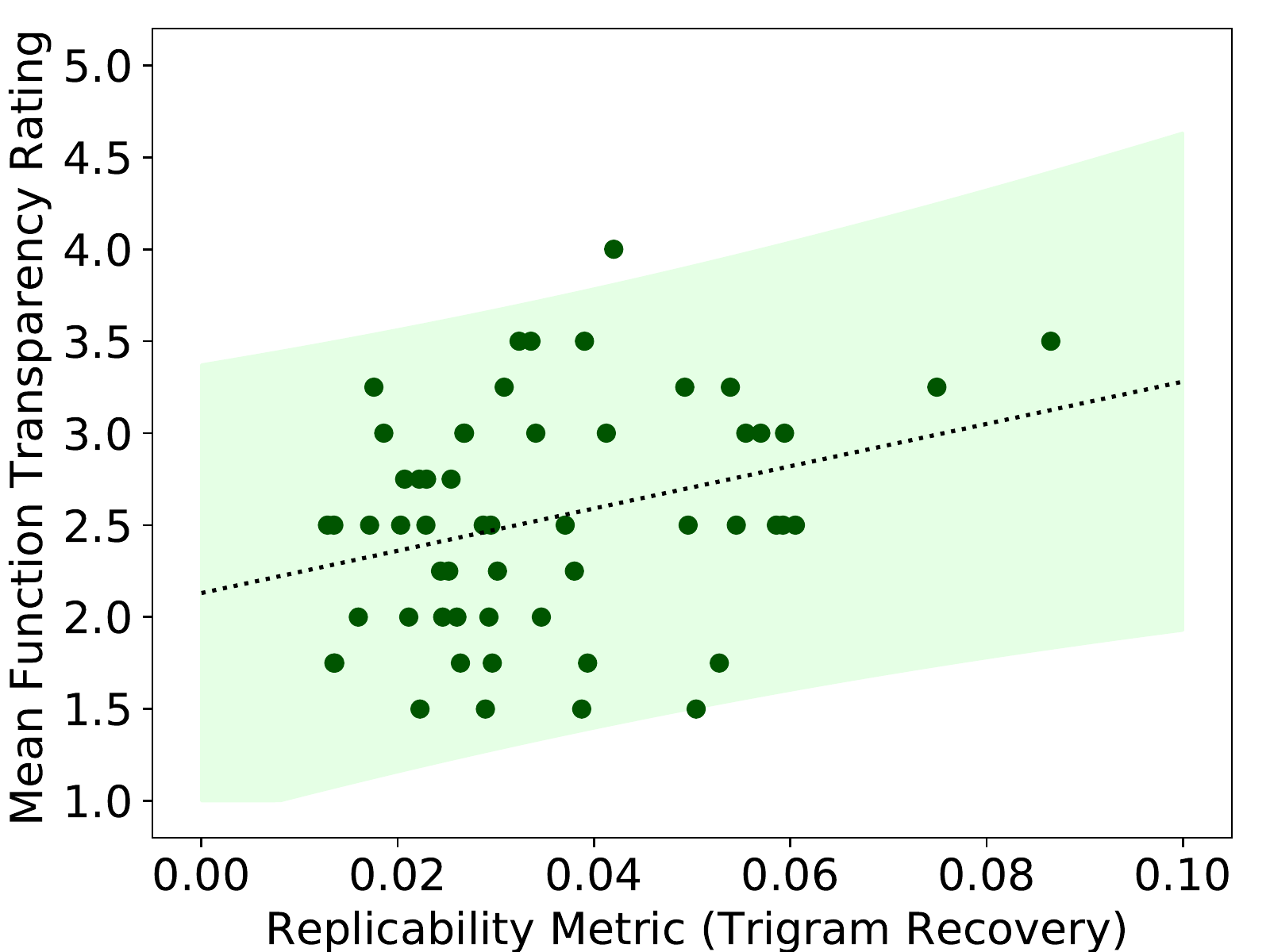}}
\end{minipage}%
\begin{minipage}{.33\linewidth}
\centering
\label{subfig:d_d}
\subfloat[$\textrm{PCC}=0.276$]{\includegraphics[width=1\linewidth]{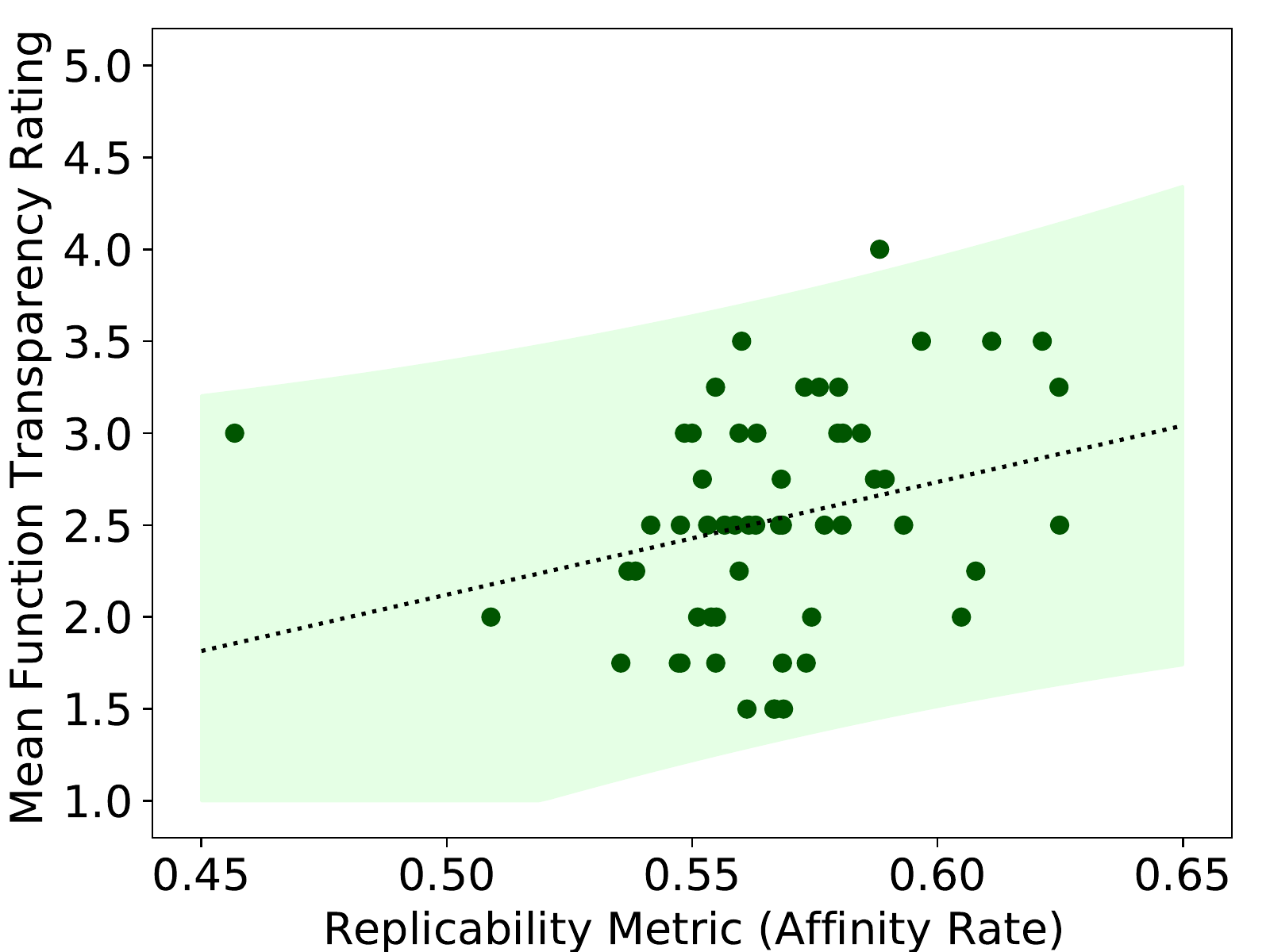}}
\end{minipage}%
\begin{minipage}{.33\linewidth}
\centering
\label{subfig:d_g}
\subfloat[$\textrm{PCC}=0.279$]{\includegraphics[width=1\linewidth]{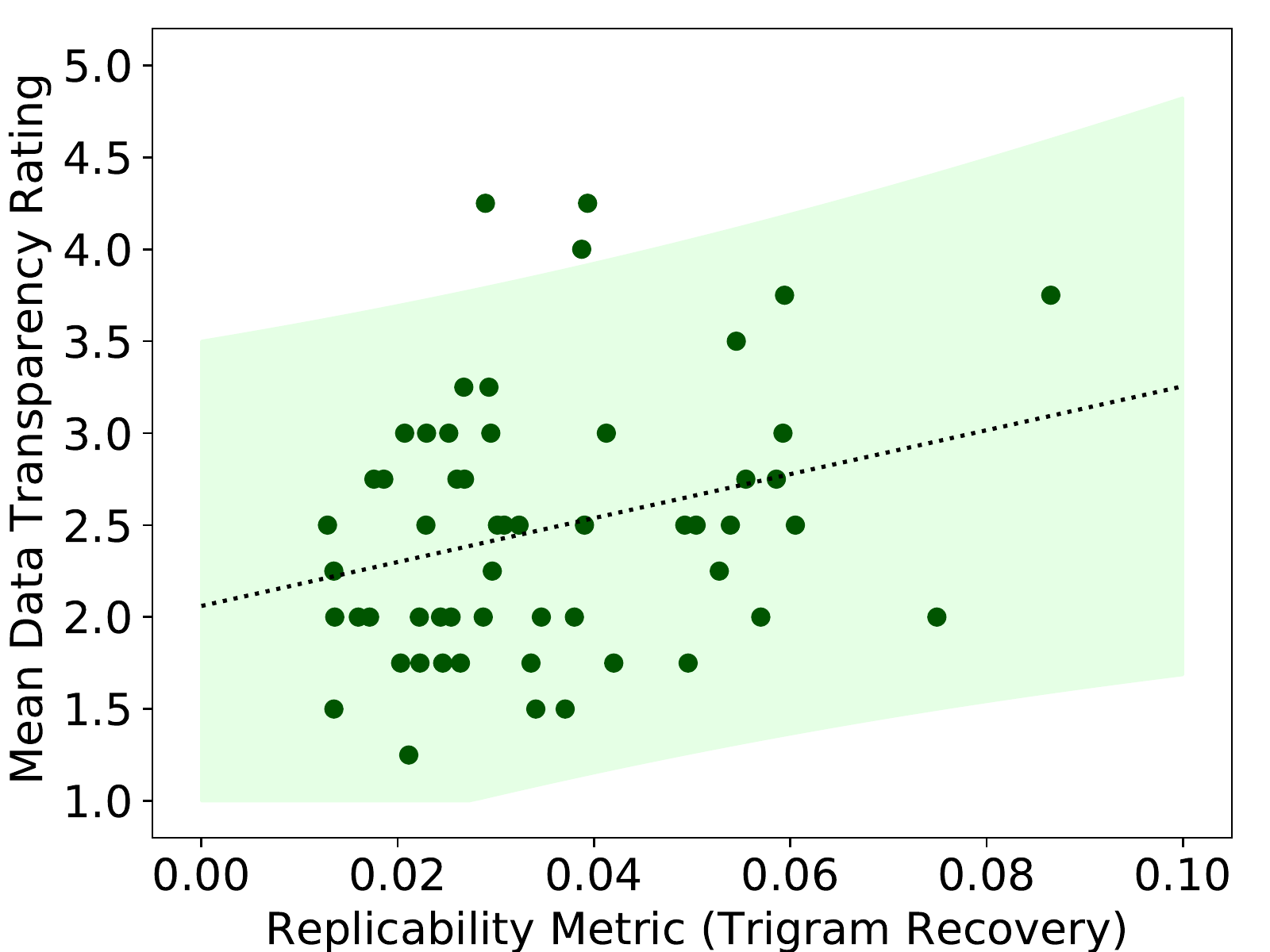}}
\end{minipage}%

\caption{The trigram recovery and sentence-affinity rate \textit{replicability} metrics behave very similarly on modeling expert function transparency opinions. Data transparency opinions are also captured by both metrics with statistically significant correlations.}
\label{fig:trigramrec}
\end{figure*}


To study the correlations both between pairs of subjective variables and between our objective metrics and the subjective responses, we use Pearson's correlation coefficient (PCC). 

In \autoref{sec:obsub} we study how the objective transparency metrics connect to expert opinion ratings along the three dimensions of transparency identified in \autoref{sec:dims}. These are the results from the first pilot study, testing if our objective metrics capture the subjective notions of disclosive transparency as understood by experts.

In \autoref{sec:response} we present the results from the second pilot study, and analyze relationships that appear between our objective transparency metrics, the subjective transparency ratings, and user opinion responses. In particular, we seek to determine if our objective measures can be used as suitable proxy for the subjective expert opinions in modeling user responses to transparency.

In this section we have scatter plots containing several overlapping $(x,y)$ coordinates. We opt to represent these points where multiple identical samples are present by varying both marker color and size to convey the relative quantities of repeats.

\subsection{Objective vs Subjective Transparency}\label{sec:obsub}

\autoref{tab:objsubj} shows the Pearson's correlations and $p$-values for the objective \textit{style-appropriateness} and \textit{replicability} transparency metrics and the subjective expert transparency ratings, and between the objective metrics themselves.

\begin{table}[t]
\small
\centering
\begin{tabularx}{0.49\textwidth}{lcccc}
\hline
\textbf{Objective Metric} & \textbf{Subj. Variable} & \textbf{PCC} & \textbf{\textit{p}}  \\
\hline
S-A    &   Task Transp.    &   0.366   &   0.006   \\
S-A    &   Function Transp.    &   -0.061  &   0.656   \\
S-A    &   Data Transp.    &   -0.030  &   0.827   \\
Replicability (TR) &   Task Transp.    &   0.150   &   0.275   \\
Replicability (TR) &   Function Transp.    &   0.309   &   0.022   \\
Replicability (TR)  &   Data Transp.    &   0.279   &   0.039   \\
Replicability (AR) &   Task Transp.    &   0.245   &   0.071   \\
Replicability (AR) &   Function Transp.    &   0.276   &   0.042   \\
Replicability (AR)  &   Data Transp.    &   -0.200   &   0.143   \\
\hline
Replicability (TR) &   S-A    &   0.200   &   0.142   \\
Replicability (AR) &   S-A    &   -0.331   &   0.014   \\
\hline\end{tabularx}
\caption{Pairwise analysis of objective style-appropriateness and replicability and expert subjective transparency scores.
}\label{tab:objsubj}
\end{table}

The style-appropriateness (S-A) metric clearly captures information about task transparency, exhibiting a positive statistically significant correlation. However, the S-A metric exhibits no significant trends with data or function transparency.

On the other hand, the replicability metric does exhibit statistically significant relationships with clear positive correlations to both function and data transparency, for trigram recovery, and with Function transparency alone for sentence affinity rate; neither exhibit statistically significant relationships with task transparency. Finally, the replicability and S-A metrics do not exhibit a significant correlation with each other.

Taken together, these results suggest that \textbf{the replicability and S-A metrics capture subjective notions of transparency}. Furthermore, they capture \textbf{complementary elements} of transparency, exhibiting no significant correlation to each other, while significantly explaining variance in different dimensions of subjective transparency.

\begin{figure*}[t]
\begin{minipage}{.33\linewidth}
\centering
\label{subfig:r_cf}
\subfloat[$\textrm{PCC}=0.347$]{\includegraphics[width=1\linewidth]{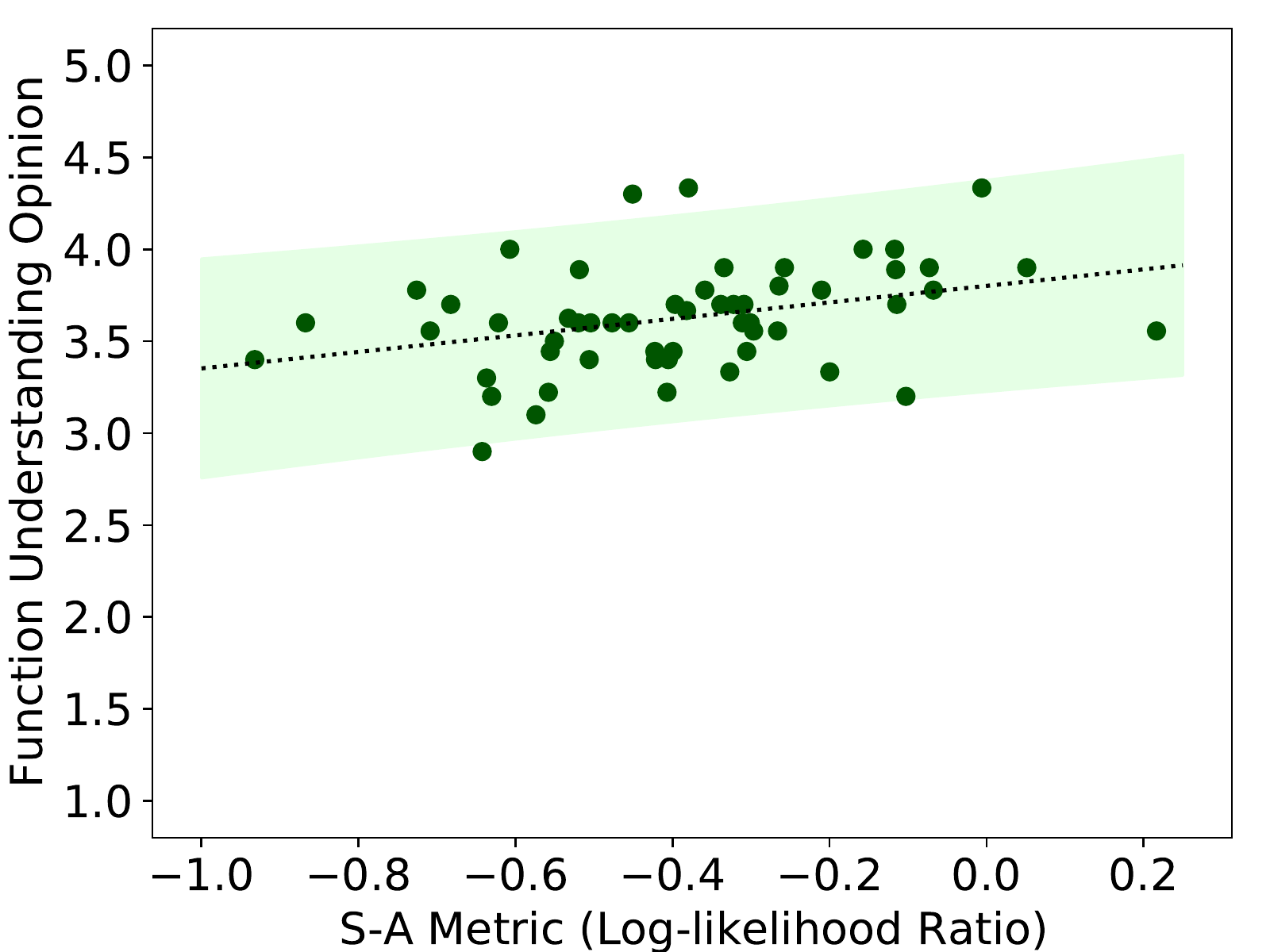}}
\end{minipage}%
\begin{minipage}{.33\linewidth}
\centering
\label{subfig:r_cg}
\subfloat[$\textrm{PCC}=0.383$]{\includegraphics[width=1\linewidth]{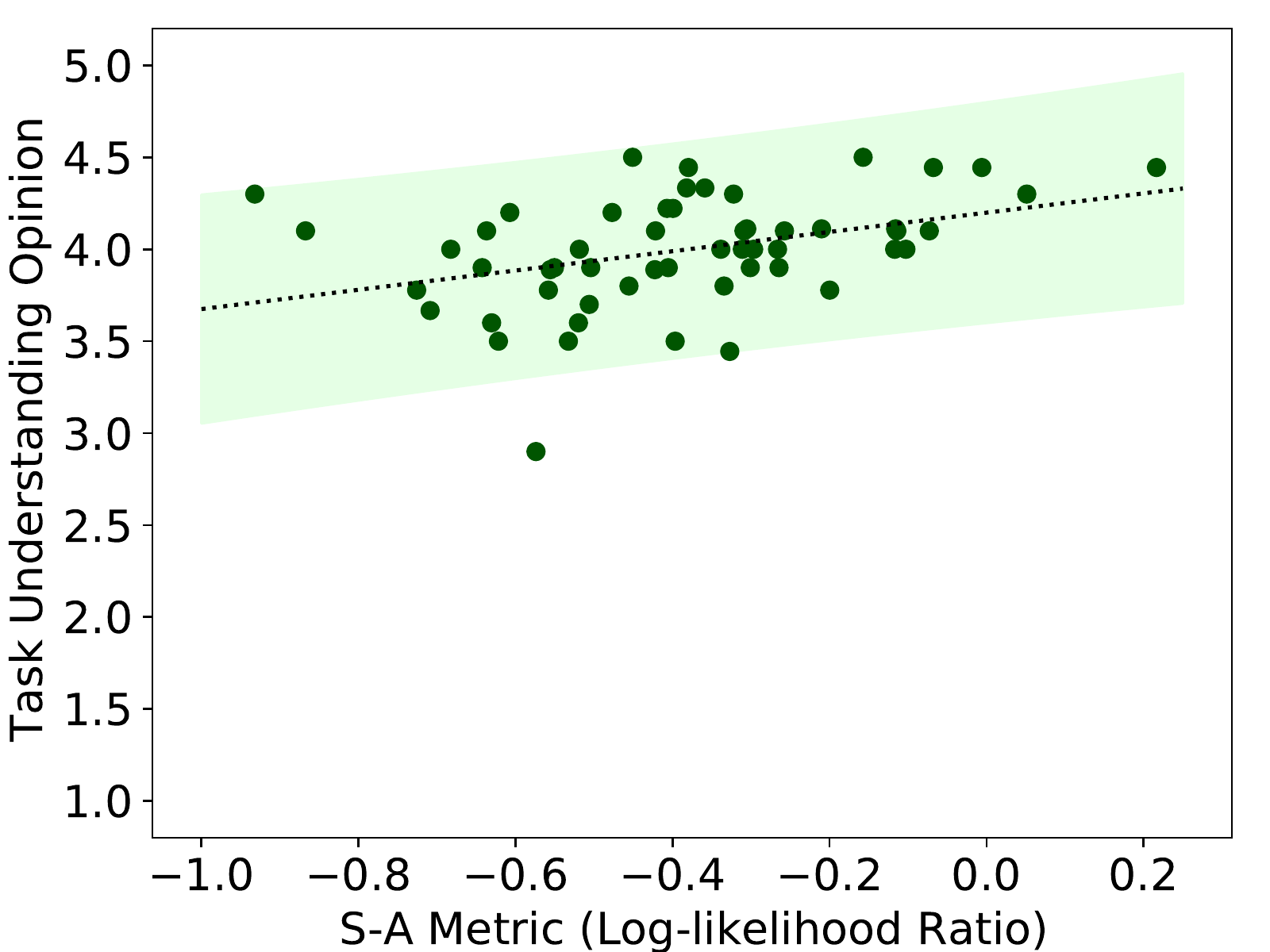}}
\end{minipage}%
\begin{minipage}{.33\linewidth}
\centering
\label{subfig:r_u}
\subfloat[$\textrm{PCC}=-0.007$]{\includegraphics[width=1\linewidth]{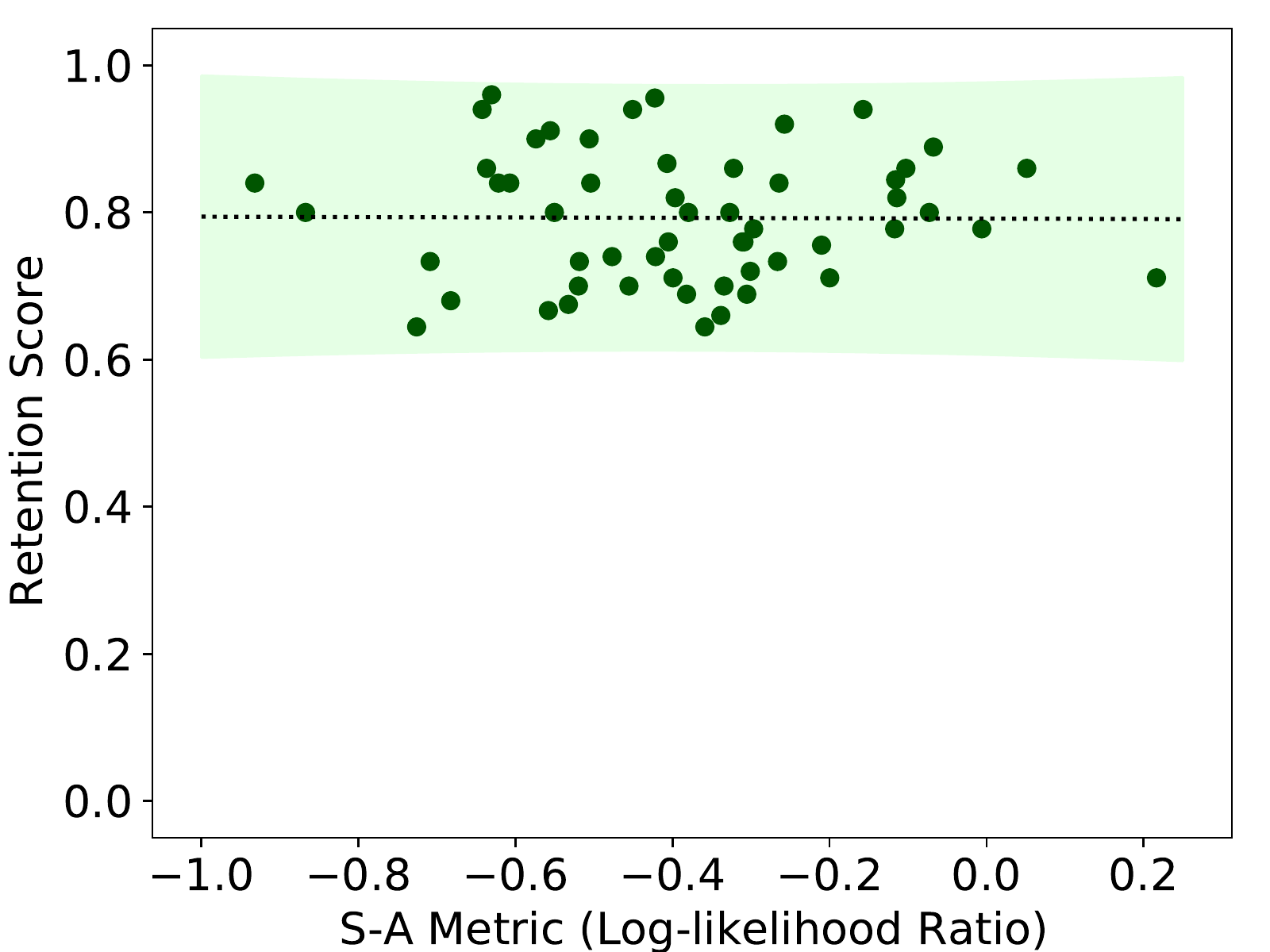}}
\end{minipage}%

\caption{The style-appropriateness log-likelihood ratio clearly captures user opinions of function and task understanding, while not capturing their real trends in retention.}
\label{fig:loglikelihood}
\end{figure*}

\autoref{fig:trigramrec} shows how our replicability trigram information recovery score is positively correlated with both expert function and data transparency ratings, however, this relationship is less strong with task transparency. Meanwhile, the S-A log-likelihood ratio score captures user opinions of both function and task understanding, but is not predictive of retention score (\autoref{fig:loglikelihood}).

While the relationship between replicability and function transparency was expected, the connection between S-A and task transparency was surprising at first. We think this is driven by a tendency for more detailed, transparent descriptions of a given task to more heavily utilize common language and analogies---after all, tasks such as translation, newsreading, and language learning all exist in the real world as topics of everyday discussion.

\begin{figure}[t]
    \centering
    \includegraphics[width=\linewidth]{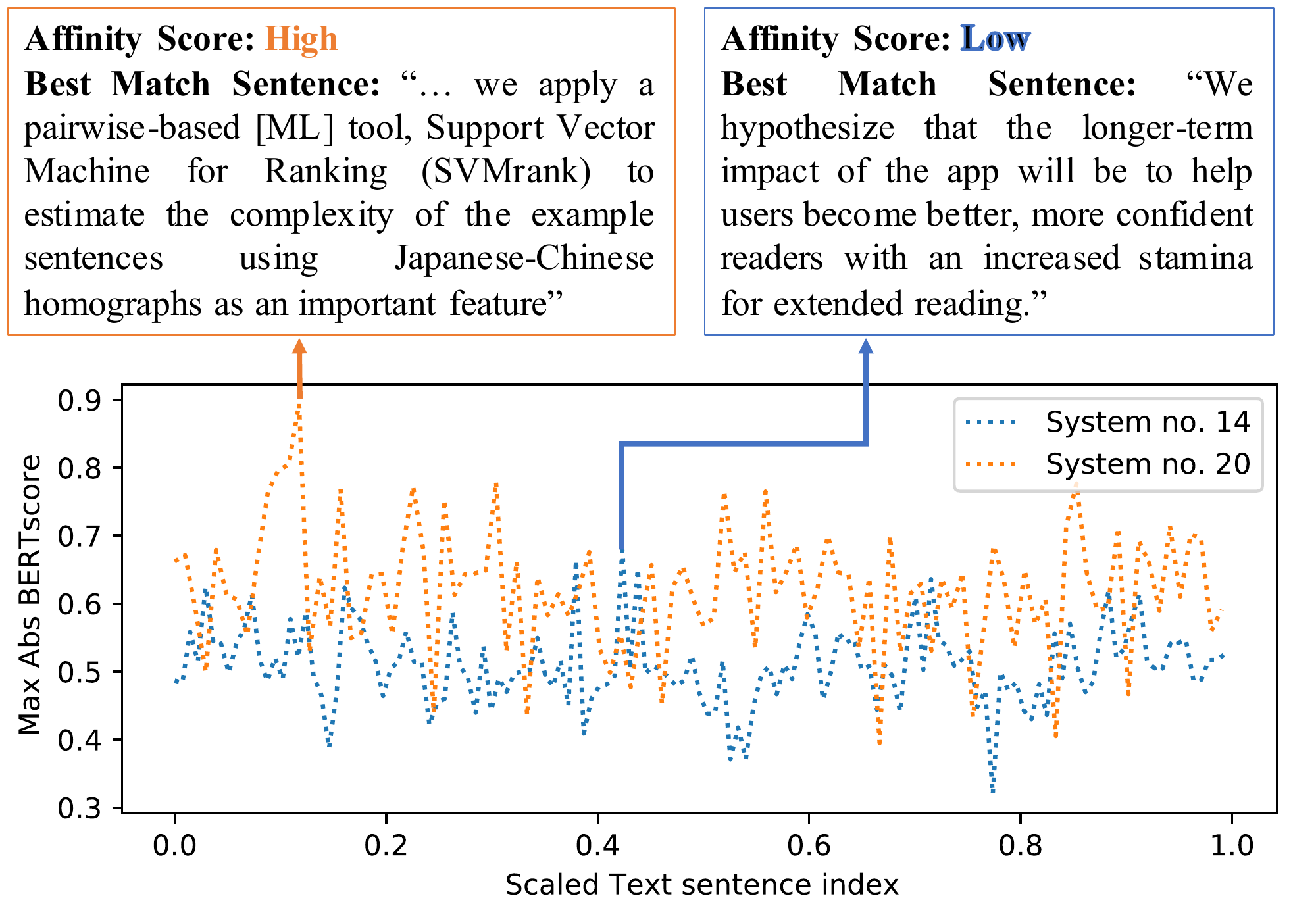} 
	\caption{Demonstrating interpretability of the BERTscore affinity replicability metric. }
	\label{fig:affhilo}
\end{figure}

Figure \ref{fig:affhilo} shows the \textit{sentence affinity rate curves} for two abstracts in the *ACL Demo Corpus. This demonstrates the interpretability of the affinity rate replicability score---the top curve has several peaks of particularly high BERTscore, which match to specific, technical sentences in the abstract providing concrete detail on how the system works. Meanwhile, the bottom curve has few high-scoring peaks, the highest of which match to one of the vague sentences in the source abstract. In other words, \textbf{descriptions with more specific technical detail in the abstract score higher on our objective transparency measures}, as desired.

\subsection{Transparency and Response Variables}\label{sec:response}

To relate the abstract-level objective and subjective transparency ratings to the response variables, we assess pairwise PCC between pairs of (transparency label, response variable) across the 55 abstracts. \autoref{tab:xy} contains a sample of pairwise correlation analyses of expert subjective transparency opinions and the user response variables. 

\begin{figure*}[t]
	\centering
\begin{minipage}{.33\linewidth}
\centering
\label{subfig:g_cf}
\centering
	\subfloat[$\textrm{PCC} =0.400$]{\includegraphics[width=1\textwidth]{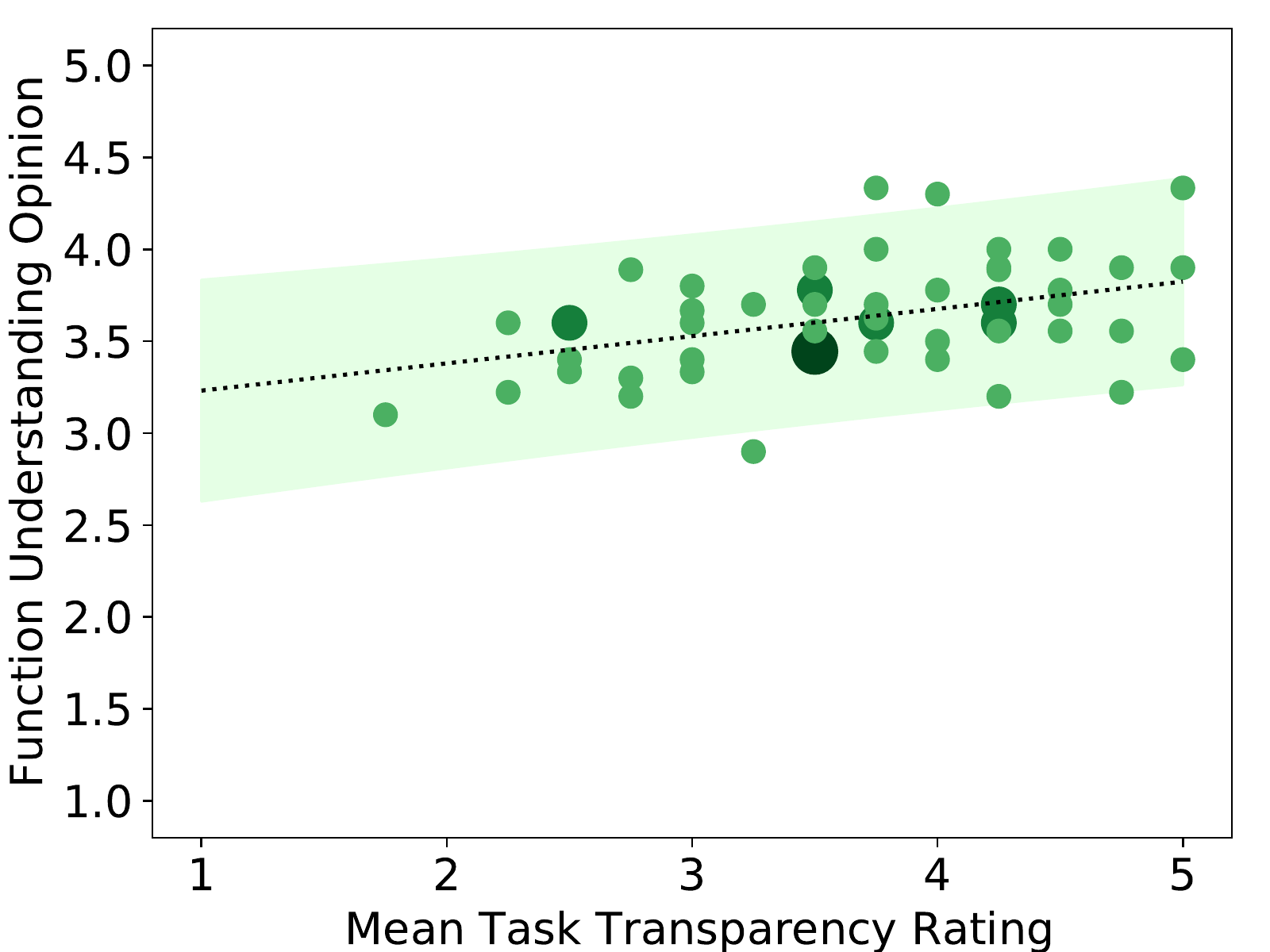}}
\end{minipage}%
\begin{minipage}{.33\linewidth}
\centering
\label{subfig:g_cg}
\subfloat[$\textrm{PCC} =0.487$]{\includegraphics[width=1\textwidth]{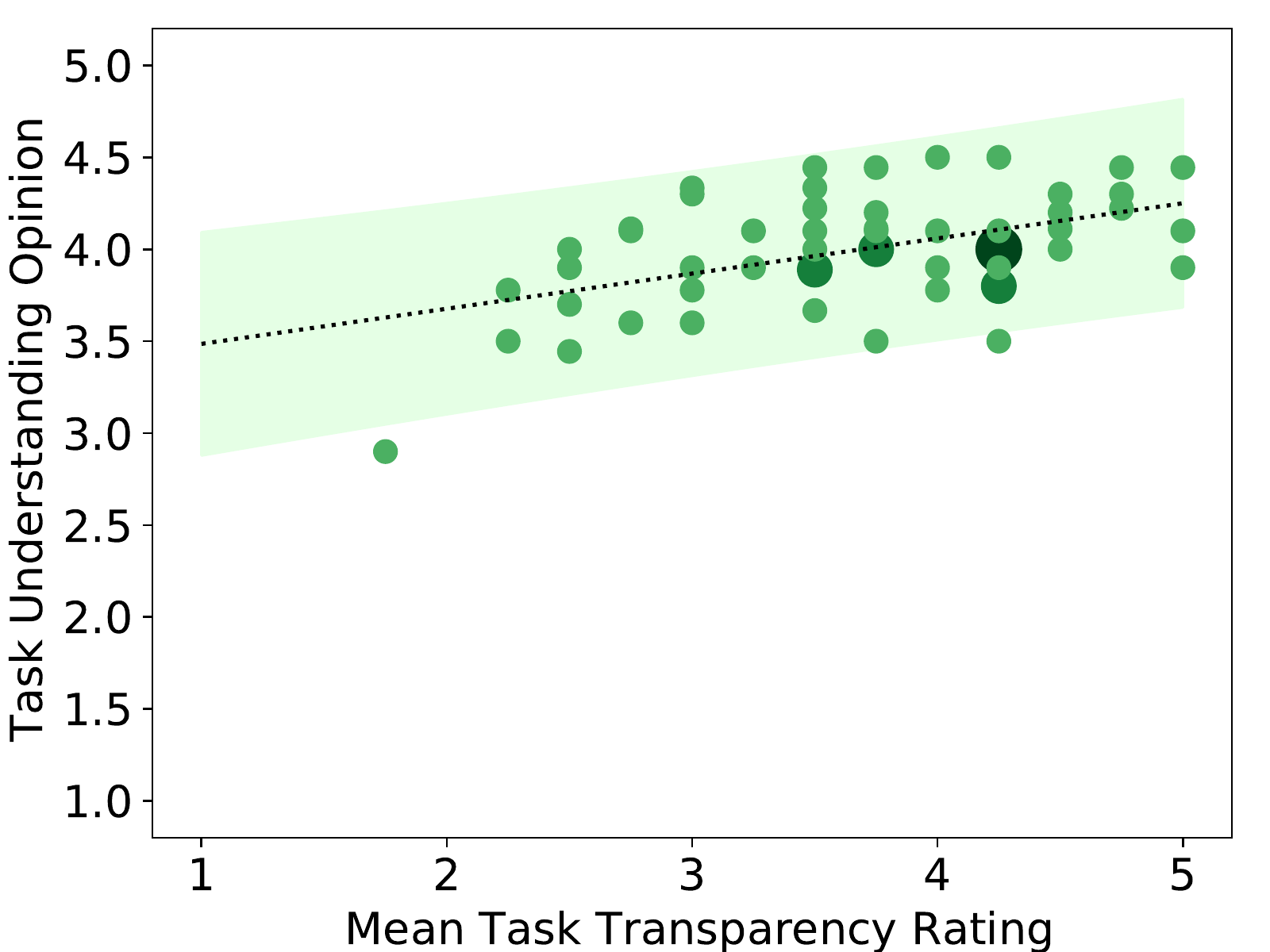}}
\end{minipage}%
\begin{minipage}{.33\linewidth}
\centering
\label{subfig:f_cf}
	\subfloat[$\textrm{PCC} =-0.157$]{\includegraphics[width=1\textwidth]{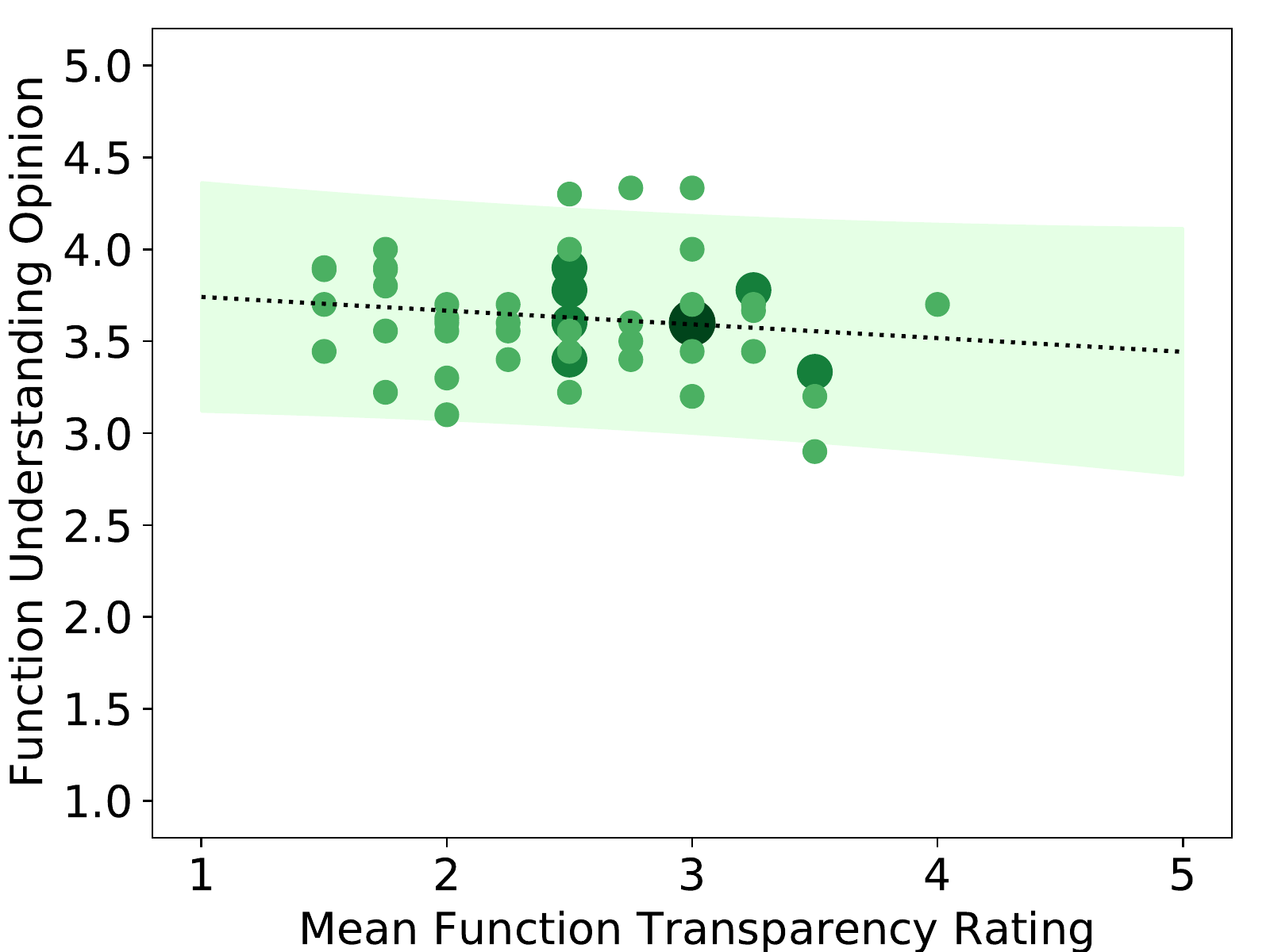}}
\end{minipage}%

	\caption{Task transparency drives a statistically significant positive trend in (a) average user function understanding and (b) average user goal understanding, but neither (c) function nor goal understanding has a trend against function transparency.}
	\label{fig:TGU}
\end{figure*}

\begin{table}[t]
\small
\centering
\begin{tabularx}{0.49\textwidth}{lcccc}
\hline
\textbf{Abstract Rating} & \textbf{Response Variable} & \textbf{PCC} & \textbf{\textit{p}}  \\
\hline
Task Transp. & Function Underst. & 0.400 & 0.003  \\
Task Transp. & Task Underst. & 0.487 & 0.000  \\
Task Transp. & Task Trust & 0.327 & 0.015  \\
Func. Transp. & Function Fairness & 0.176 & 0.199  \\
Funct. Transp. & Task Fairness & -0.236 & 0.083  \\
Data Transp. & Function Underst. & 0.292 & 0.030  \\
\hline\end{tabularx}
\caption{Pairwise correlations between abstract-averaged subjective transparency scores and user response variables.}\label{tab:xy}
\end{table}

Of the statistically significant correlations between subjective scores we observe, the positive relationship between expert-rated average \textbf{task transparency} and average user self-reported \textbf{task understanding} is the strongest. Similarly, there is a strong, statistically significant positive correlation between \textbf{task transparency} and self-reported \textbf{function understanding}. However, we were surprised to find that our expert-rated average \textbf{function transparency} does not exhibit a statistically significant correlation with either of the aforementioned understanding variables. \autoref{fig:TGU} depicts the positive correlations of these two response variables with task transparency.

In other words, users believe they understand not only the task, but the function of the system being described better, the more well-motivated and transparent the discussion of the \textit{task} is. However, transparency in describing the function of the system will not lead to users to think they understand either aspect better. This result was surprising.

As function transparency captures ``how the system works,'' while task transparency is concerned with ``what the system does,'' it is probably the case that understanding the task is a necessary prerequisite for understanding the system function. This could explain the connection between task transparency and user function understanding. However, this connection is illusory: for example, a detailed description of the problem of translating French to Arabic contains no information about the attention mechanisms used by the underlying transformer.


\begin{table}[t]
\small
\centering
\begin{tabularx}{0.49\textwidth}{lcccc}
\hline
\textbf{Objective Metric} & \textbf{Response Var.} & \textbf{PCC} & \textbf{\textit{p}}  \\
\hline
S-A    &   Func. Underst.   &   0.347   &   0.009   \\
S-A    &   Task Underst.   &   0.383   &   0.004   \\
Replicability (TR) &   Func. Fairness   &   0.285   &   0.035   \\
Replicability (TR) &   Task Fairness   &   -0.351  &   0.009   \\
Replicability (AR) & Task Fairness & -0.258 & 0.0573 \\
Replicability (TR) &   Data Trust  &   0.242   &   0.075   \\
Replicability (TR) &   Task Trust  &   0.286   &   0.034   \\
\hline\end{tabularx}
\caption{Pairwise correlations objective transparency metrics and user response variables.}\label{tab:autoxy}
\end{table}

\section{Related Work}
Most previous work on explainability and intelligibility in transparency focuses on ``explanations that are contrastive and sequential rather than purely subjective'' \cite{miller2019explanation}. Furthermore, prior studies regarding transparency have tended to focus on transparency in post-facto user judgments of automated systems following their use. For example, \citet{kizilcec_how_2016} and \citet{wang_factors_2020} both studied how \textit{system users} viewed the fairness of a system that judged them. In both studies, the authors found that while transparency affects perceived fairness, the strongest predictor of perceived fairness was final outcome favorability. 


Work in explainable AI (XAI) intersects with disclosive transparency when asking what kinds of explanations are preferred by users. Several studies have directly asked users of experimental XAI systems to respond to  \cite{bhatt2019explainable, kaur2020interpreting, hong2020human} the quality and effects of the explanations. Judging how well system function descriptions are motivated by explainable outputs is thus a promising direction.

Many solutions intended to remedy ethical concerns of machine learning include a disclosive component. \citet{lim2009and} studied how novice end-users can understand the function in context of intelligible systems. Similarly to us, they derive explanations from fundamental questions of ``what,'' ``why,'' and ``how'' that resolve the gulf of evaluation and gulf of execution in the user, as put forward by \cite{norman1988psychology}.

While we are primarily concerned with disclosure to laypeople, disclosure between practitioners has been a key thrust of work in recent years. Disclosing potential deficiencies, biases, failure points, and intended use cases for datasets \cite{holland2018dataset, bender2018data, gebru2018datasheets}, pretrained models \cite{mitchell2019model}, and full systems \cite{arnold2019factsheets} is crucial in ensuring the ethical construction and deployment of systems that utilize machine learning while minimizing the risk of potential harms. All of these proposals focus heavily on outlining best practices to ensure that various sources of bias are accounted for and made available to system builders and decision makers before harms are perpetuated.

\citet{lipton2018mythos} introduced \textit{simulatability}, the capacity for a user to comprehend all constituent operations of a model by manually performing them, as fundamental to transparency. This relies on a notion of \textit{decomposability} \cite{lou2012intelligible}.
 Intelligibility is both an end in itself and a useful tool for ensuring buy-in from stakeholders such as users \cite{muir1994trust} and management \cite{veale_2018_fairness}. 
 \citet{knowles2017intelligibility} deal with designing systems that use intelligibility to convey evidence of trustworthiness under uncertain scenario, and frame intelligibility as the degree to which the system's underlying logic can be conveyed to users. 
 




\section{Conclusions}

Our replication-based framework allows us to tackle the problem of characterizing disclosive transparency in NLP system descriptions. We developed \textit{style-appropriateness} and \textit{replicability} metrics using neural language models, which capture multiple subjective dimensions of transparency. We ran a pilot study characterizing layperson responses to varying levels of transparency in system descriptions, demonstrating the value of the conceptual framework and the concrete tools proposed. We release all code, data, and annotations online at \texttt{ \href{https://github.com/michaelsaxon/disclosive-transparency}{https://github.com/michaelsaxon/}} \texttt{\href{https://github.com/michaelsaxon/disclosive-transparency}{disclosive-transparency/}}.

Three natural directions for future work follow these results. The automated transparency metrics can remove the costly expert annotator constraint, enabling scaled up transparency-user opinion relationship studies. More sophisticated conditional abstract-to-full text inversion models can be directly swapped in with the replicability metric to produce better automated transparency scores. Finally, the metrics can be used to guide NLG processes, perhaps to enable transparency-constrained abstractive summarization or style transfer.




\section{Acknowledgements}

This work was supported in part by the National Science Foundation Graduate Research Fellowship under Grant No. 1650114. We would also like to thank the Robert N. Noyce Trust for their generous gift to University of California via the Noyce Initiative. The views and conclusions contained in this document are those of the authors and should not be interpreted as representing the sponsors.

\section{Ethical Considerations}

\autoref{tab:autoxy} demonstrates that increased \textbf{replicability} scores are associated with decreasing opinions of the ethics of the underlying task (increasing disagreement with the statement ``I believe this system is made to solve an ethical task'')  but increasing opinions of function fairness (increasing disagreement with ``I am concerned about the fairness of this system''). This was a surprising result that we are interested in attempting to replicate on larger corpora. With this relatively small sample size, it is hard to interpret why these conflicting trends in ethical opinions might arise. 

This negative relationship between replicability and task ethics could be an artifact of the data used to train the abstract-to-full text inversion model underpinning the replicability metric. It might be the case that there is higher public awareness and controversy toward ML tasks that are more well-represented in the arXiv inversion dataset. However, this result might also represent a genuine ethical quandary for further research in transparency.

Much of Section 1 focused on making the case for \textit{why measuring transparency is important}. However, if it really is the case that increased measurable disclosive transparency drives decreasing trust in system ethics, it's conceivable that system providers or governments could become less inclined to disclose. We would consider this a negative outcome. However our results so far do not overwhelmingly suggest this relationship exists.


Another potential limitation of this study is that layperson end-users are not the target audience of these NLP conference abstracts. The decision to use academic writings as stimuli was driven by the availability of (short description, long description) pairs consisting of the abstracts and their corresponding full texts. However, we believe the short description/long document pairing is applicable to a number of fields, including terms of service as discussed in Section 2, and to other types of systems described in non-academic, but still technical text, such as patent disclosures and instruction manuals.

Finally, as attitudes toward privacy, fairness, and ethics can be culturally variable, we note that our survey results were sourced from crowd workers in majority-English speaking countries and may not be broadly representative.




\bibliographystyle{acl_natbib}
\bibliography{anthology,acl2021}



\appendix

\section{The Replication Room}


Many insights that drove the aforementioned metrics sprang from a thought experiment inspired by \citet{searle1980minds} called a ``replication room.'' 

Imagine a room in which a person sits with access to a hypothetically comprehensive set of domain-specific and general resources on machine learning (ML), natural language processing (NLP), software engineering, programming languages, which collectively enable them to look up any applicable term of art or find instructions on implementing any standard component or system. In this replication room, this \textit{implementer} receives system descriptions $e$ from which they are tasked with producing a functionally analogous reproduction of the system $i$ being described.

Under this scenario, clearly no system description can be less transparent than a blank piece or paper or completely unrelated text. In any of these edge scenarios, the description provides no clues about the system the implementer is intended to reproduce, and their chance of success is nothing but the prior over all possible implementable systems. Thus, in a channel communication sense no information is conveyed.  

On the other hand, a system description from which every detail is accounted for would represent a maximum transparency scenario. A complete printout of all elements of the source code and all setup/training steps, or a detailed tutorial, or even the full text of an academic system description paper could be sufficient for a successful reproduction. In these cases, no further information content would increase the level of transparency, as all information needed is included.

We dub the degree to which a system description contains the necessary information to perform a replication its \textit{replicability.} Replicability as a notion of transparency is incomplete, however, as it makes no assumptions about the degree to which the supporting materials might be used, or the kind of person who is performing the replication. 

Another component is the \textit{style-appropriateness} of the description. In short, the style of language in the description, the overall quality of writing and ease of reading, the level of reliance on unexplained technical jargon, and various other factors all impact the degree to which a layperson would require use of the assistive materials, or the level of expertise that would be required for a replication to be successfully carried out without using the assistive materials. More precisely, the style-appropriateness of the description is tied to the degree to which domain knowledge about the AI problem area is necessary to reconstruct $i$ from the conventional meaning $s$ in \citet{bender-koller-2020-climbing}'s framework. 

\section{Details from Crowd Worker Survey}\label{app:surveydeets}

Each crowd worker was shown a survey containing a system description abstract with instructions to provide their level of agreement on a five-point Likert opinion scale (strongly disagree, disagree, neutral, agree, strongly agree) to a set of six prompts intended to assess their level of ``task understanding,'' ``task fairness,'' ``function understanding,'' ``function fairness,'' ``function trust,'' and ``data trust.'' The prompts provided for each opinion value are provided in \autoref{tab:userprompts}. After the users completed the six opinion questions they were instructed to press a button to hide the abstract, after which the retention questions were revealed. Each subject was compensated with $\$0.10$ USD per survey, and subjects averaged a Human Intelligence Task (HIT) competition time of 35 seconds. This translates to an average hourly compensation of \$10.29 per hour. 

\subsection{Opinion Prompts}

\begin{table}[ht!]
\centering
\small
\begin{tabularx}{0.5\textwidth}{l  X}
\hline
\textbf{Opinion Value} & \textbf{Prompt} \\
\hline
\textbf{Task Understanding} & I understand what this system is meant to do. \\
\textbf{Task Fairness} & I believe this system is made to solve an ethical task. \\
\textbf{Function Understanding} & I understand how this system works. \\
\textbf{Function Fairness} & I am concerned about the fairness of this system. \\
\textbf{Function Trust} & I think this system can accurately perform its task. \\
\textbf{Data Trust} & I think this system will protect my privacy and data. \\

\hline
\end{tabularx}

\caption{The prompts provided to assess the six user opinion values. Crowd workers provide five-point scale agreement opinions to each statement.}\label{tab:userprompts}
\end{table}

All of the above opinion prompts---except for function fairness---produce outputs where ``strongly agree'' corresponds to a positive position, such as confidence in one's own understanding or in the performance of the system. Thus, we regularize the reverse the polarity of the function fairness scores by reversing the order of numerical assignment to responses, assigning 5 to ``strongly disagree'' with being concerned about fairness, 4 to ``disagree,'' etc.

While we are only interested in subjective user opinions about trust and fairness, for understanding we seek to analyze the ``truth,'' as users might be overconfident in their pure opinions. To do this we must develop some feasible objective measure---for this study we use retention as a proxy for understanding.

\subsection{Retention Evaluation}

Out of a desire for some objective assessment of user confusion, we additionally ask participants to recall whether a set of phrases did or did not appear in the abstract they just read. The simple metric of retention accuracy reveals both how carefully a participant read the abstract and how much they maintained it in their memory---while these questions fail to directly capture user understanding in the way that a conceptual quiz would, they can be generated automatically and consistently across topic areas. 


After completing the opinion question section of the survey, crowd workers are instructed to press a ``Hide Passage'' button. Once the button is pressed, the system description abstract is greyed out, and a set of five retention questions is revealed. Each abstract has its own set of retention questions, which are randomly generated prior by sampling sentences either from the abstract or from the other abstracts in the dataset.

\section{Supplementary Analysis of User Responses}\label{app:suppl}

\subsection{User Opinion Agreement}

Although the opinion scores are subjective, and we don't necessarily expect agreement, some measure of agreement in opinions between users on abstracts is desirable to further support the validity of our abstract-wise correlation analysis. We are unable to use PCC as a measure of inter-rater reliability here because no two abstracts were rated by the same 10 crowd workers. Thus, we are restricted to analyzing the average within-abstract variance of each response variable. \autoref{tab:respirr} contains the average abstract-wise variance for each of the opinion response variables.

\begin{table}[t]
\centering
\small
\begin{tabular}{lc}
\hline
\textbf{Response Variable} & \textbf{Avg. Abstract-wise Variance} \\

\hline
Func. Undst. & 1.17 \\
Task Undst. & 0.700 \\
Func. Fair. & 1.25 \\
Task Fairness & 1.22 \\
Data Trust & 0.73 \\
Task Trust & 0.71 \\
\hline
\end{tabular}
\caption{User agreement for the user responses using assessed average per-abstract response variance for the subjective metrics.}\label{tab:respirr}
\end{table}

\subsection{Task Domain as a Confounder}

\begin{figure}[t]
    \centering
    \includegraphics[width=0.48\textwidth]{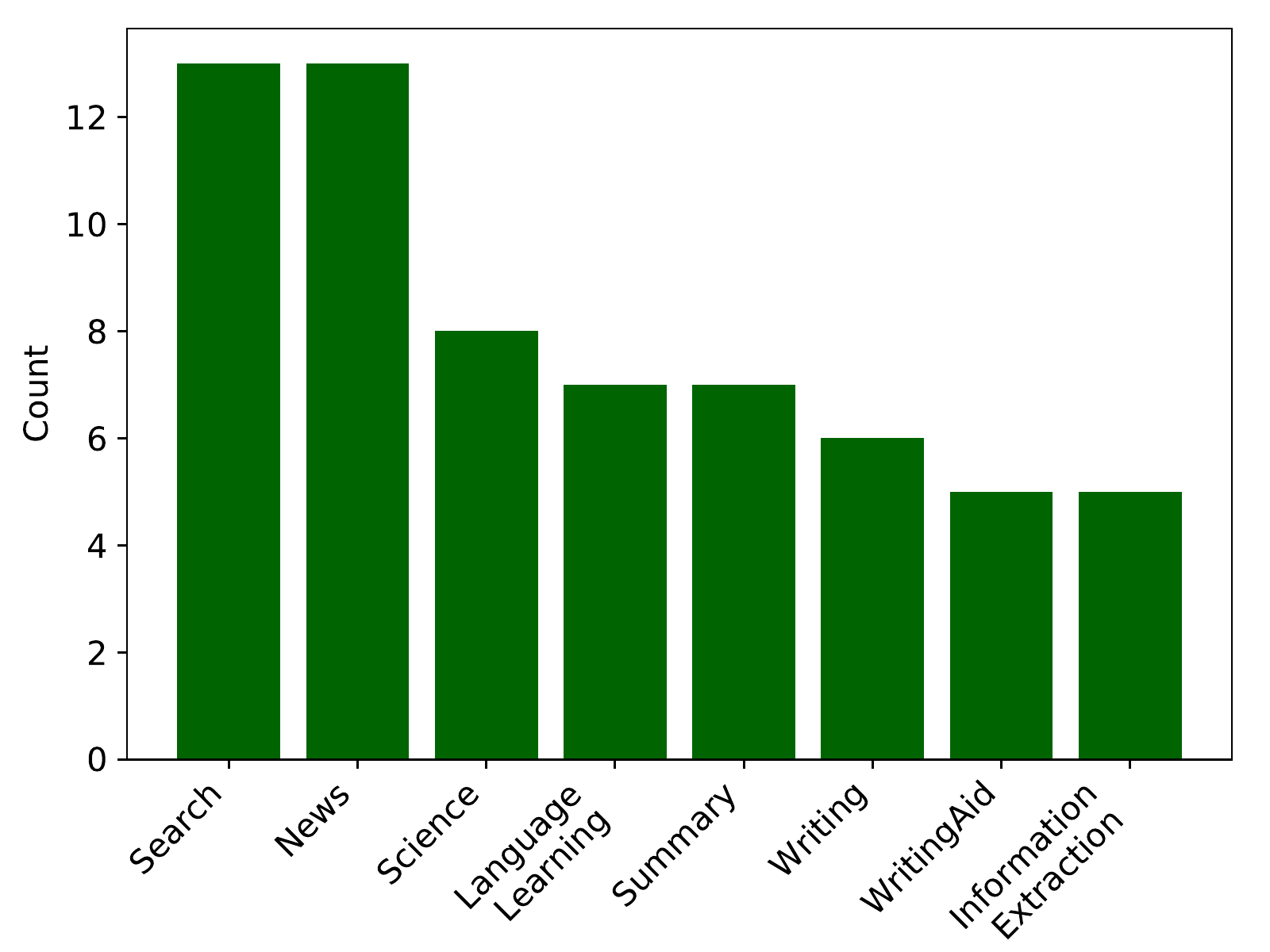}
    \caption{Distribution of topical keywords possessed by at least five abstracts.}
    \label{fig:kwbar}
\end{figure}

For each abstract we collect topical keywords to handle the potential confound of differing user opinions by problem domain, to enable analysis for whether the broad problem area a system is intended to solve (e.g., newsreaders, language learning apps, translation tools) is a confounder driving positive or negative user attitudes. \autoref{fig:kwbar} depicts the top keywords in our dataset.

Because we are evaluating user attitudes toward a diverse set of systems in distinct task domains, differences in popular perceptions toward the various tasks might bias user opinions. To ascertain if this is the case, we perform pairwise Student's \textit{t} significance tests across all pairs of keywords present in 5 or more different abstracts, on each of the subjective response variables. \autoref{tab:ttest} contains the pairs of keywords that have statistically significant differences between their distributions and the response variables along which these differences occur.

\begin{table}[t]
\small
\centering
\begin{tabularx}{0.5\textwidth}{lcccc}
\hline
\textbf{Response} & \textbf{Keyword 1} & \textbf{Keyword 2} & \textbf{\textit{t}} & \textbf{\textit{p}} \\
\hline
Func. Fair. & News & LangLearn & -3.268 & 0.005 \\
Func. Fair.& Science & LangLearn & -2.231 & 0.043 \\
Task Fair. & News & Search & 2.338 & 0.029 \\
Task Fair. & News & Writing & 2.472 & 0.032 \\
Data Trust & News & Science & -2.307 & 0.035 \\
Data Trust & Science & Search & 2.807 & 0.011 \\
Task Trust & Science & Writing & 2.368 & 0.037 \\
\hline
\end{tabularx}
\caption{Pairwise Student's t-test statistics for keyword domain classes of abstracts.}\label{tab:ttest}
\end{table}

\begin{figure}[t]
	\centering
	\includegraphics[width=0.5\textwidth]{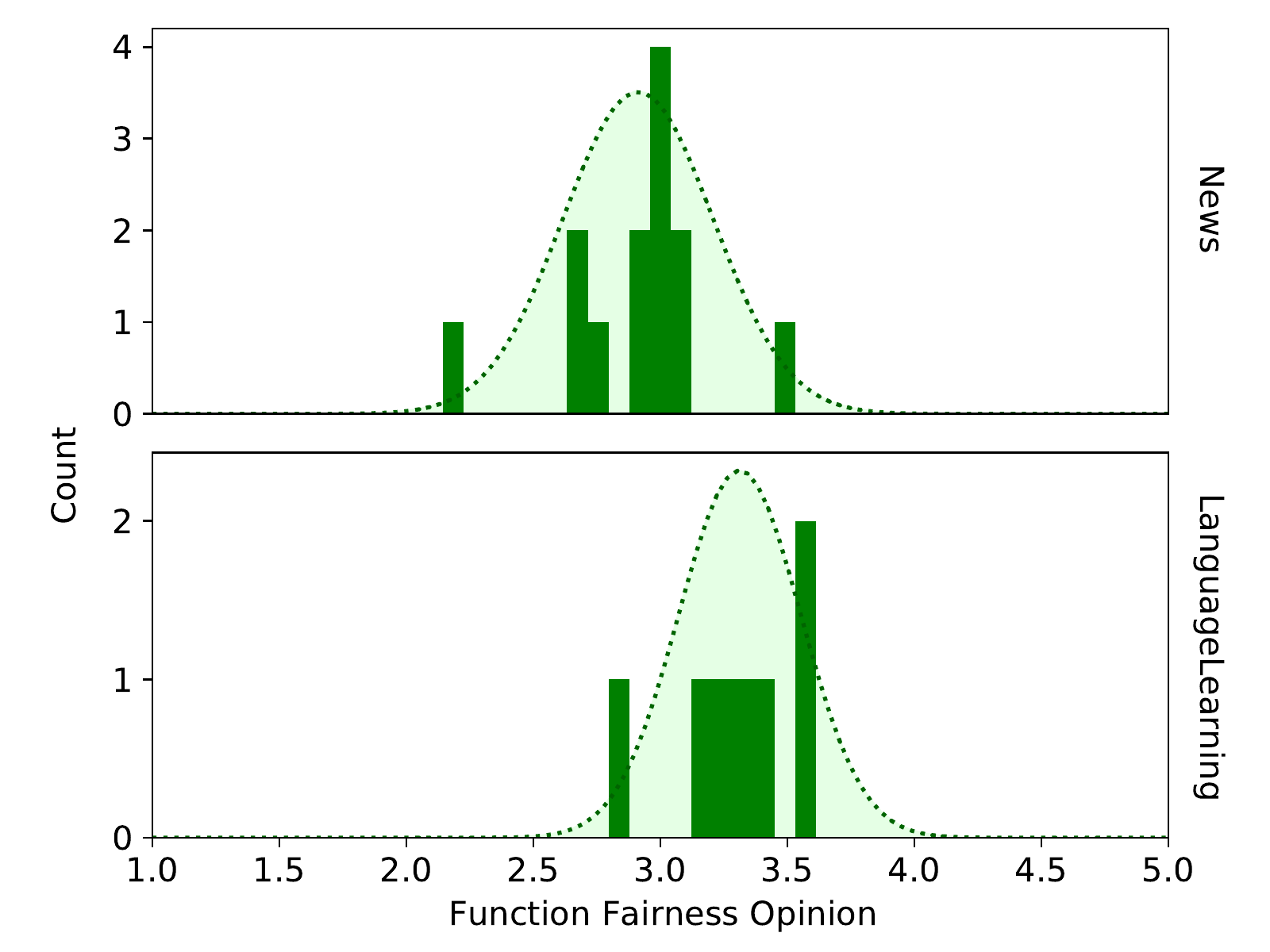}
	\caption{Separation of mean function fairness opinion distributions between news- and language learning-related abstracts, ($t=-3.3$, $p=0.005$).}
	\label{fig:NewsLL}
\end{figure}

Of these, the most significant difference was between \textit{news} and \textit{language learning} abstracts on function fairness. \autoref{fig:NewsLL} depicts the distributions of the two sets on function fairness. It shows that the users tend to believe systems developed to aid in language learning are less prone to unfair bias than news reading, summarization, and synthesis applications. This difference could be driven by genuine differences in attitudes toward the components, subtasks, and processes used to build these systems, but there's a chance it is driven by societal attitudes toward learning and news instead.

Unfortunately, these pairwise keyword attitude difference tests are severely limited by small sample sizes. It is likely that more interesting analysis could be performed on a larger dataset.

\section{arXiv Dataset Preparation}\label{app:arx}

After sampling \texttt{.tex} files from the aforementioned ranges, they were converted into training data according to the following procedure:

\begin{enumerate}
    \item If the manuscript is stored across multiple \texttt{.tex} files, collate them into a single one by replacing all \texttt{\textbackslash{}input} and \texttt{\textbackslash{}insert} commands with their corresponding file text.
    \item Split the resulting file using the \texttt{\textbackslash{}begin\{\}} and \texttt{\textbackslash{}end\{\}} tags for \texttt{abstract} and \texttt{document}. 
    \item Using the \texttt{pylatexenc} python package\footnote{\url{https://pylatexenc.readthedocs.io/en/}} use the \texttt{latex\_to\_text()} command to convert the abstract and document latex code into unicode.
    \item Remove all redundant whitespace, tabs, and new line characters, and convert all resultant text to lower case.
\end{enumerate}

\section{Regarding use of PCC}

We use PCC rather than rank-correlation coefficients such as Spearman's because we are interested in evaluating whether our metrics can implicitly capture relative deltas opinions in order to be a useful proxy for the opinions themselves in future work. 


\begin{figure*}[t]
    \centering
    \begin{mdframed}
    \textbf{Opinion Questions}
    \\\\
    \textit{Please provide your agreement with the following questions on a five-point scale from ``I strongly agree'' to ``I strongly disagree.''}
    \\\\
    \textbf{I understand what this system is meant to do.} \\
    \begin{tabular}{ccccc}
    $\bigcirc$ Strongly Disagree & $\bigcirc$ Disagree & $\bigcirc$ Neutral & $\bigcirc$ Agree & $\bigcirc$ Strongly Agree
    \end{tabular}
        \\\\
    \textbf{I understand how this system works.} \\
    \begin{tabular}{ccccc}
    $\bigcirc$ Strongly Disagree & $\bigcirc$ Disagree & $\bigcirc$ Neutral & $\bigcirc$ Agree & $\bigcirc$ Strongly Agree
    \end{tabular}
    \\\\
    \textbf{I think this system can accurately perform its task.} \\
    \begin{tabular}{ccccc}
    $\bigcirc$ Strongly Disagree & $\bigcirc$ Disagree & $\bigcirc$ Neutral & $\bigcirc$ Agree & $\bigcirc$ Strongly Agree
    \end{tabular}
        \\\\
    \textbf{I think this system will protect my privacy and data.} \\
    \begin{tabular}{ccccc}
    $\bigcirc$ Strongly Disagree & $\bigcirc$ Disagree & $\bigcirc$ Neutral & $\bigcirc$ Agree & $\bigcirc$ Strongly Agree
    \end{tabular}
    \\\\
    \textbf{I am concerned about the fairness of this system.} \\
    \begin{tabular}{ccccc}
    $\bigcirc$ Strongly Disagree & $\bigcirc$ Disagree & $\bigcirc$ Neutral & $\bigcirc$ Agree & $\bigcirc$ Strongly Agree
    \end{tabular}
        \\\\
    \textbf{I believe this system is made to solve an ethical task.} \\
    \begin{tabular}{ccccc}
    $\bigcirc$ Strongly Disagree & $\bigcirc$ Disagree & $\bigcirc$ Neutral & $\bigcirc$ Agree & $\bigcirc$ Strongly Agree
    \end{tabular}
    \\\\
    \textbf{After you have completed the opinion questions, press ``Hide Passage'' to reveal the retention questions.}
    \\\\
    \begin{tikzpicture}
    \draw  (0,0) rectangle (3,1.5) node[pos=.5] {\textbf{Hide Passage}};
    \end{tikzpicture}

    \end{mdframed}  \caption{The opinion questions section of the survey shown to Mechanical Turk crowdworkers.}
    \label{fig:survey}
\end{figure*}


\begin{figure*}[t]
	\centering
\begin{minipage}{.33\linewidth}
\centering
\centering
\includegraphics[width=1\textwidth]{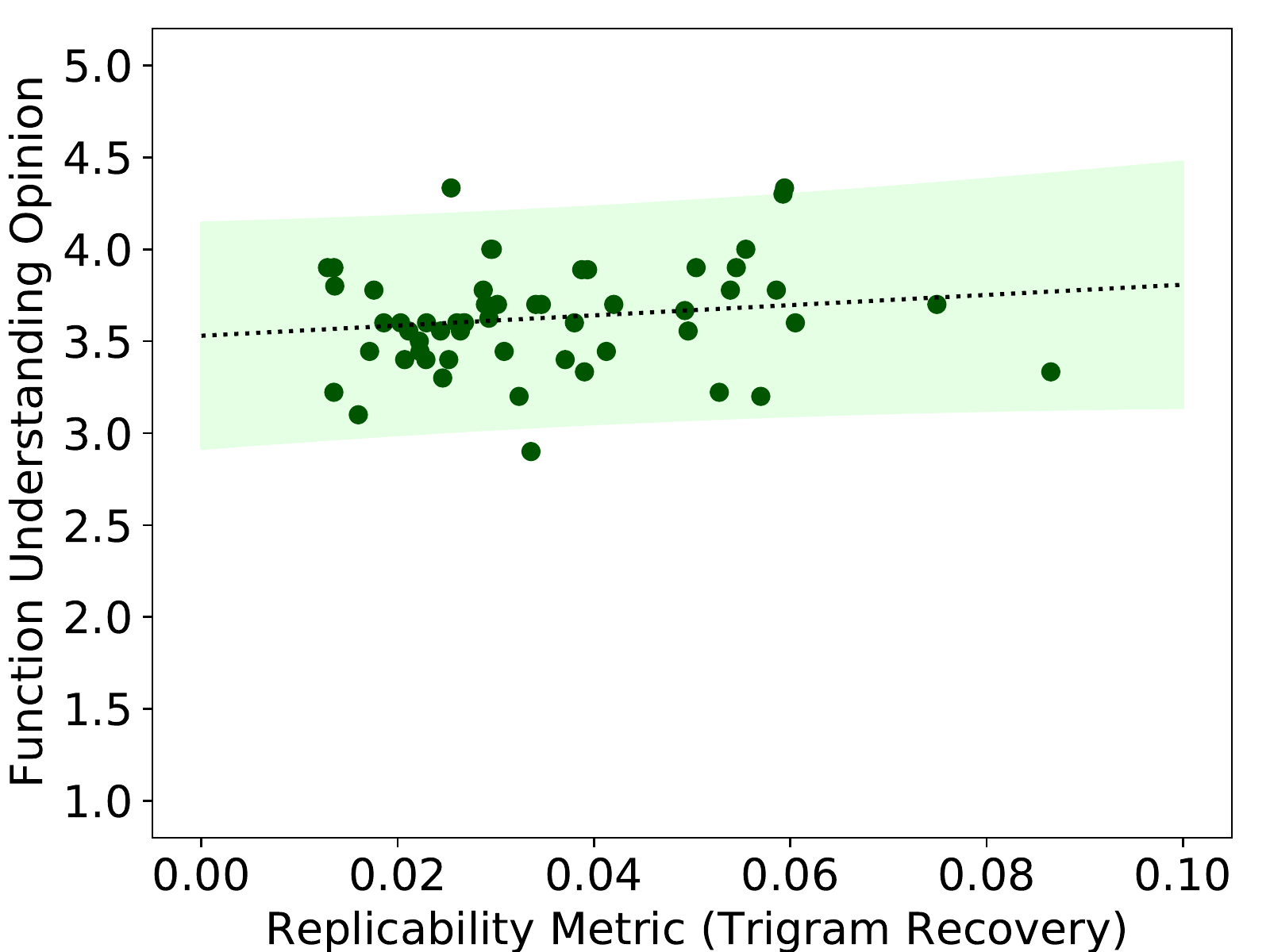}
\end{minipage}%
\begin{minipage}{.33\linewidth}
\centering
\includegraphics[width=1\textwidth]{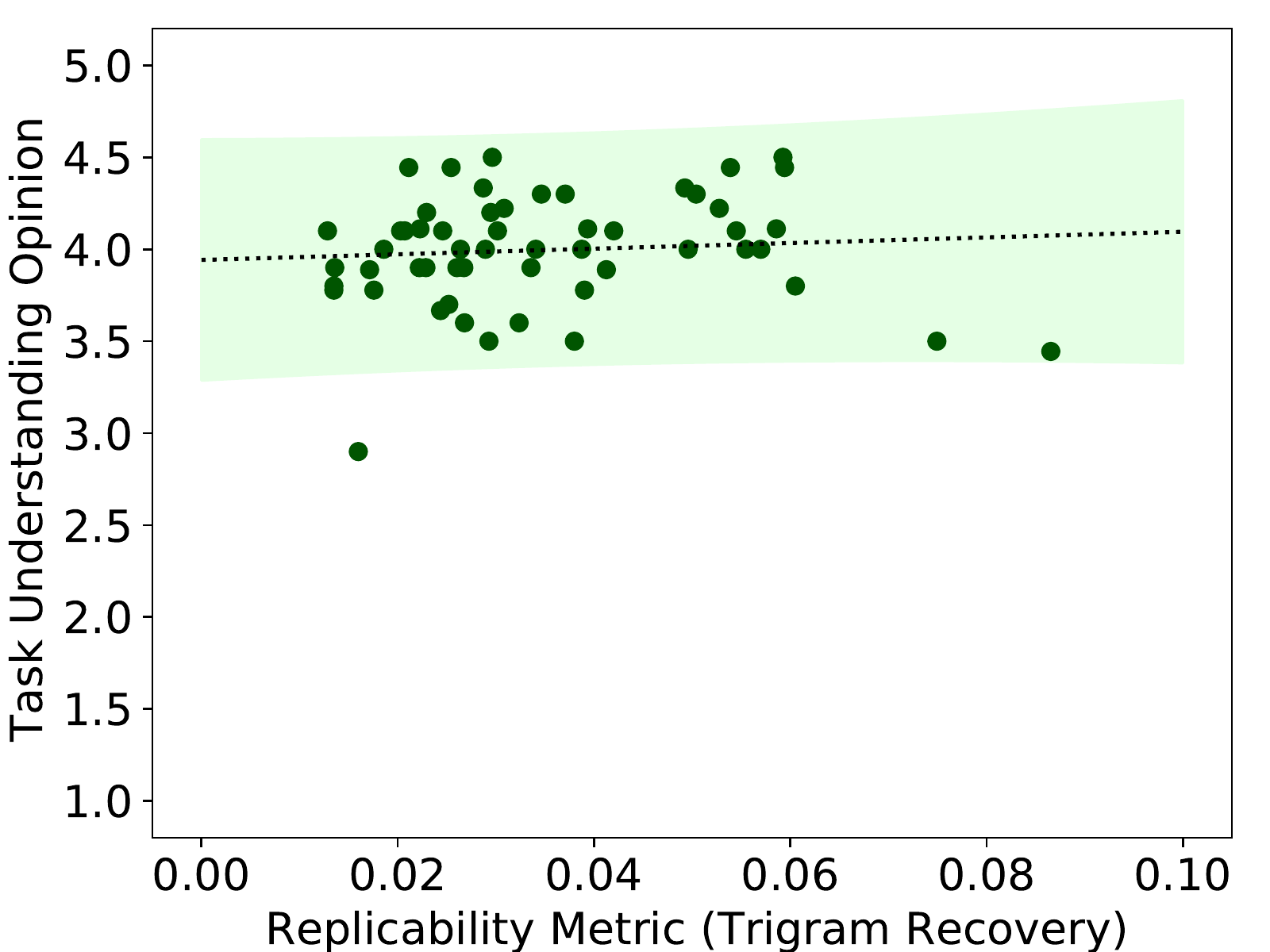}
\end{minipage}%
\begin{minipage}{.33\linewidth}
\centering
\includegraphics[width=1\textwidth]{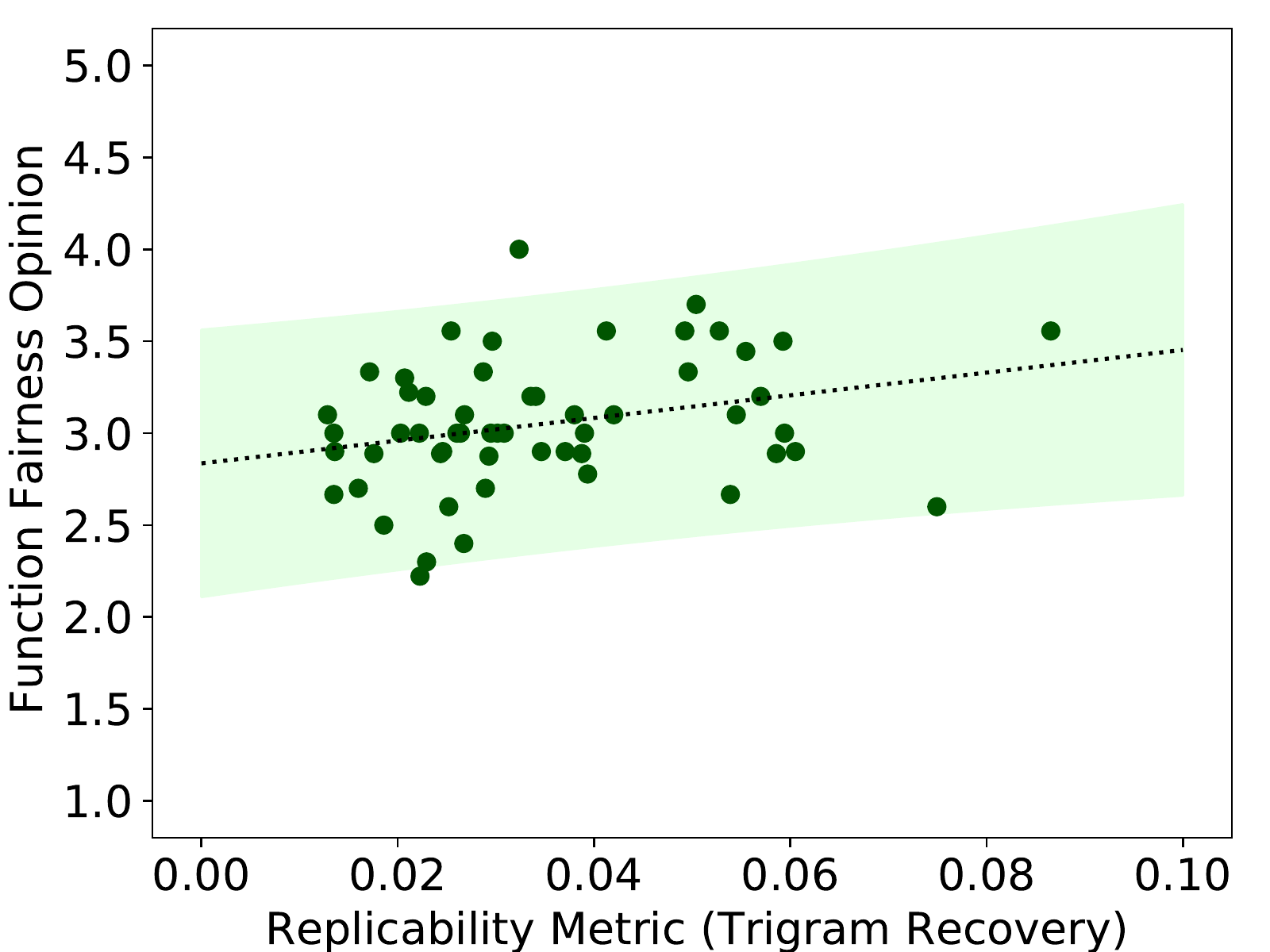}
\end{minipage}\\

\begin{minipage}{.33\linewidth}
\centering
\centering
\includegraphics[width=1\textwidth]{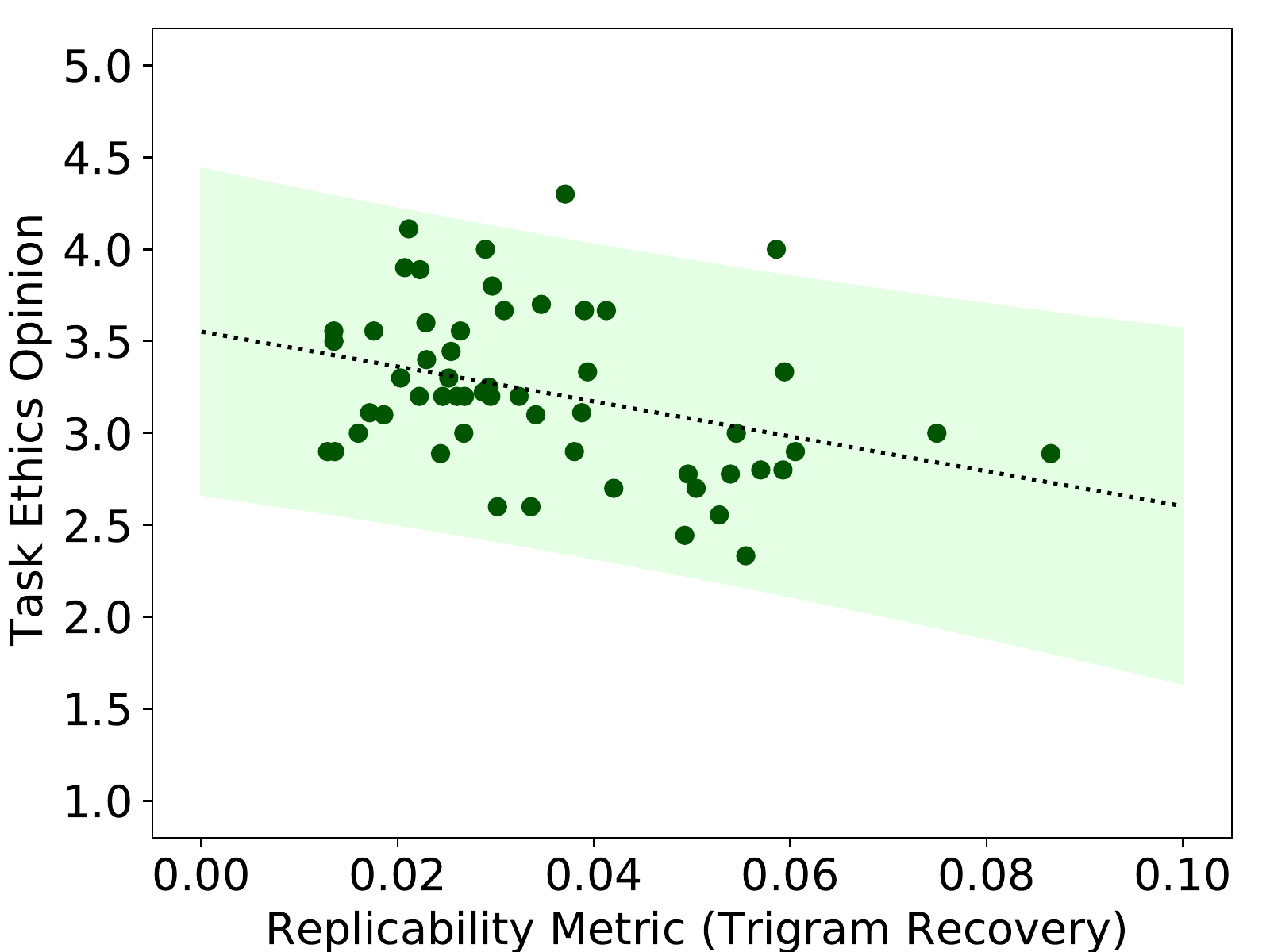}
\end{minipage}%
\begin{minipage}{.33\linewidth}
\centering
\includegraphics[width=1\textwidth]{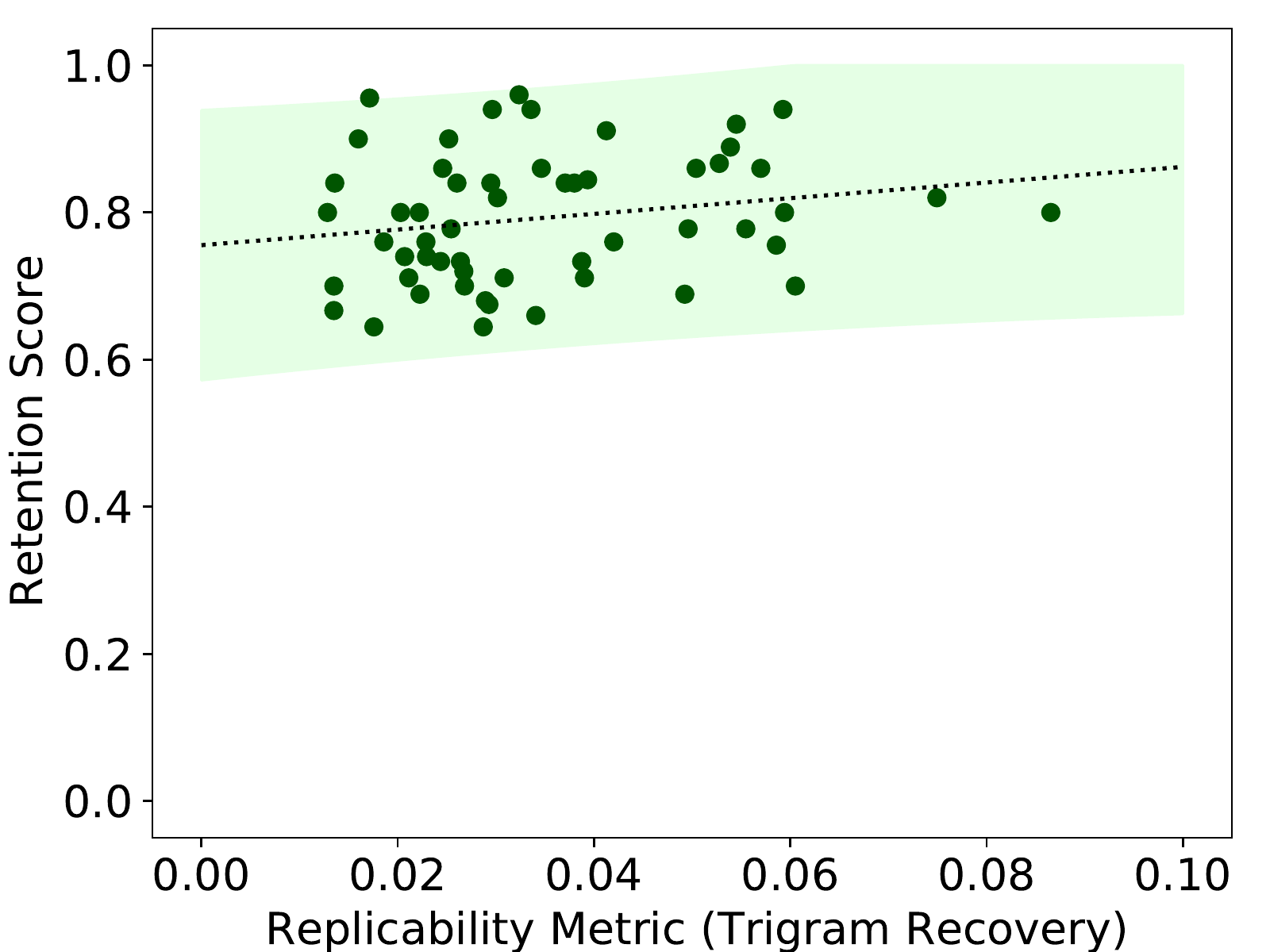}
\end{minipage}%
\begin{minipage}{.33\linewidth}
\centering
\includegraphics[width=1\textwidth]{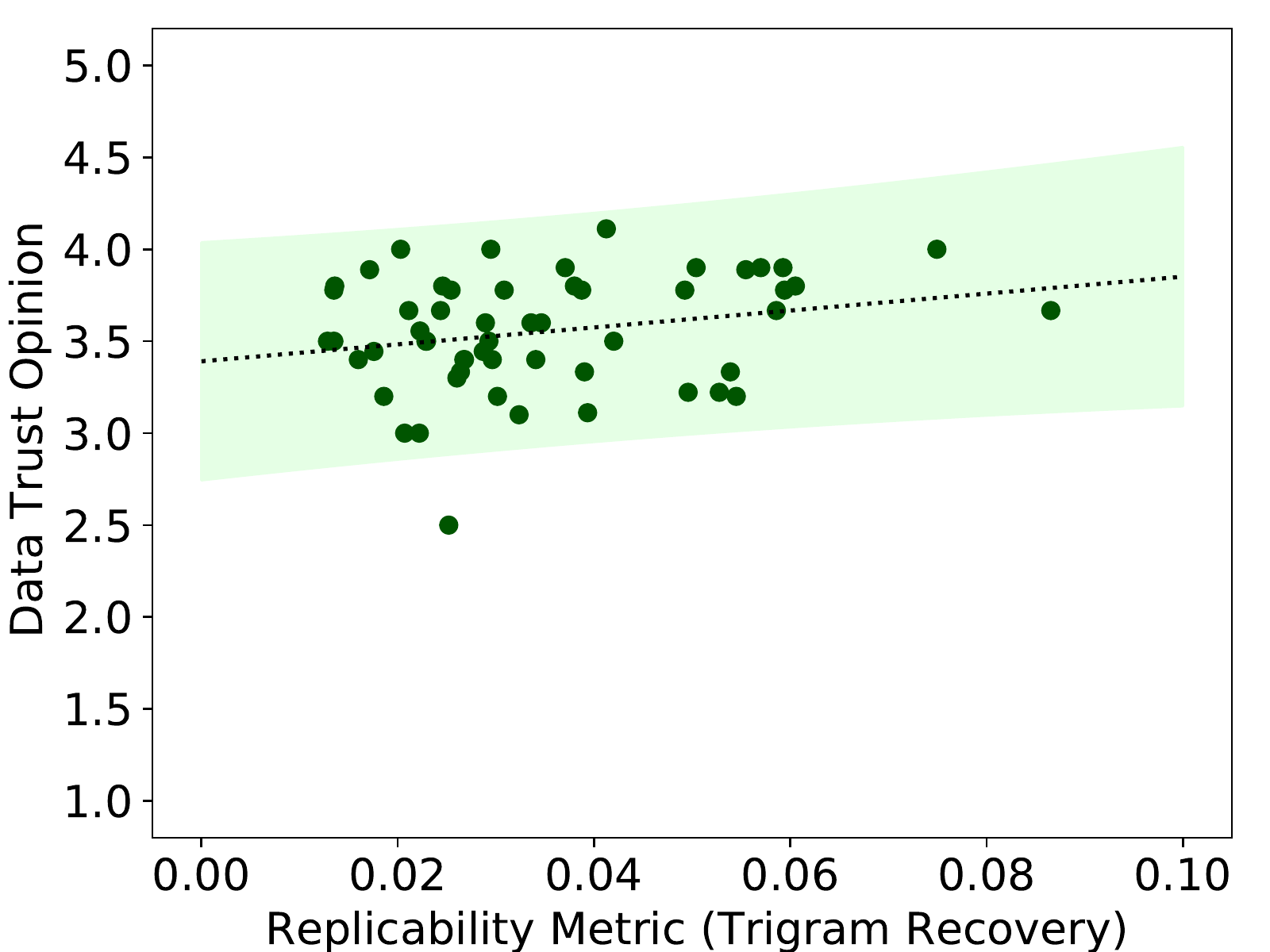}
\end{minipage}\\

\begin{minipage}{.33\linewidth}
\centering
\centering
\includegraphics[width=1\textwidth]{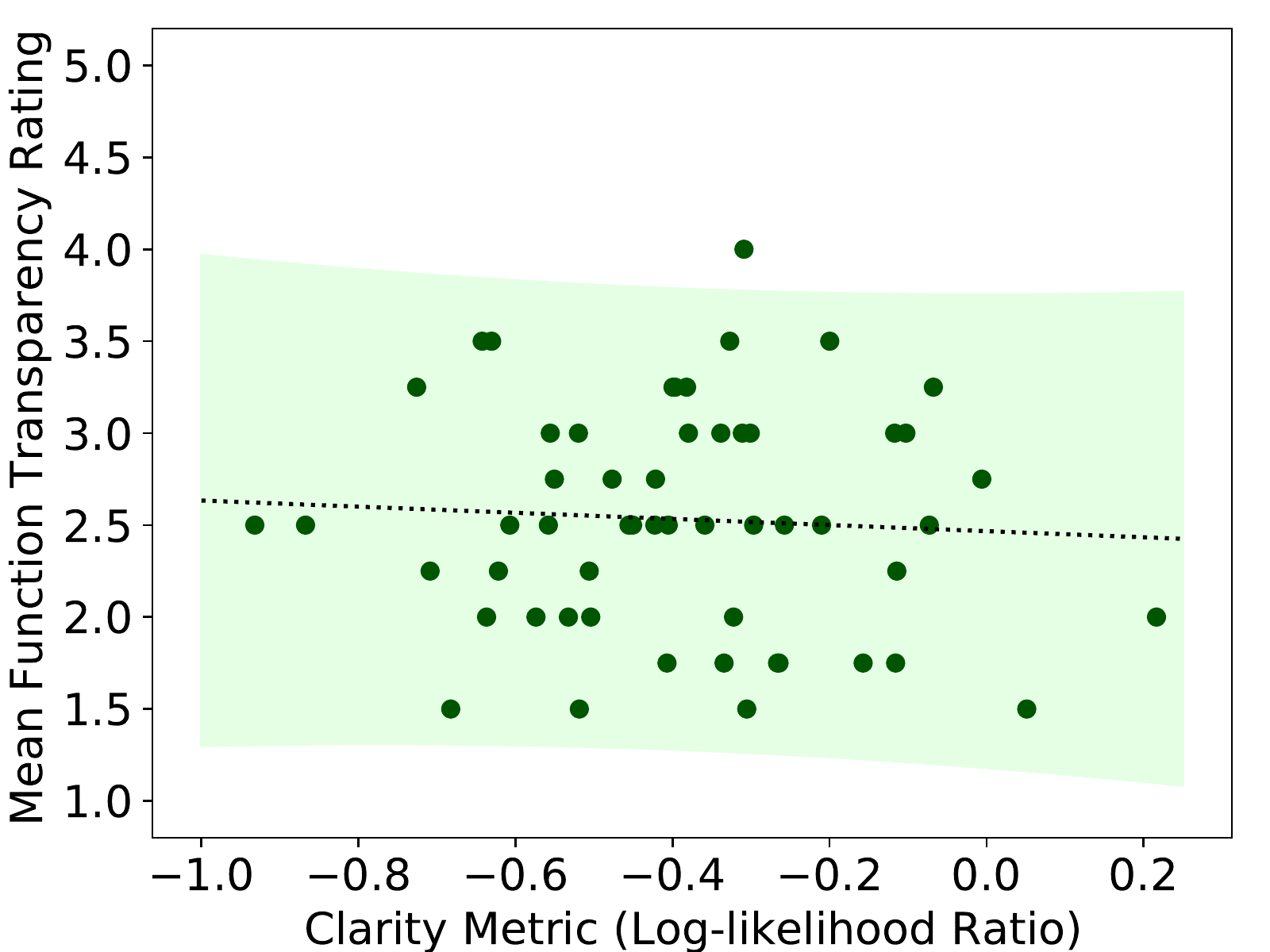}
\end{minipage}%
\begin{minipage}{.33\linewidth}
\centering
\includegraphics[width=1\textwidth]{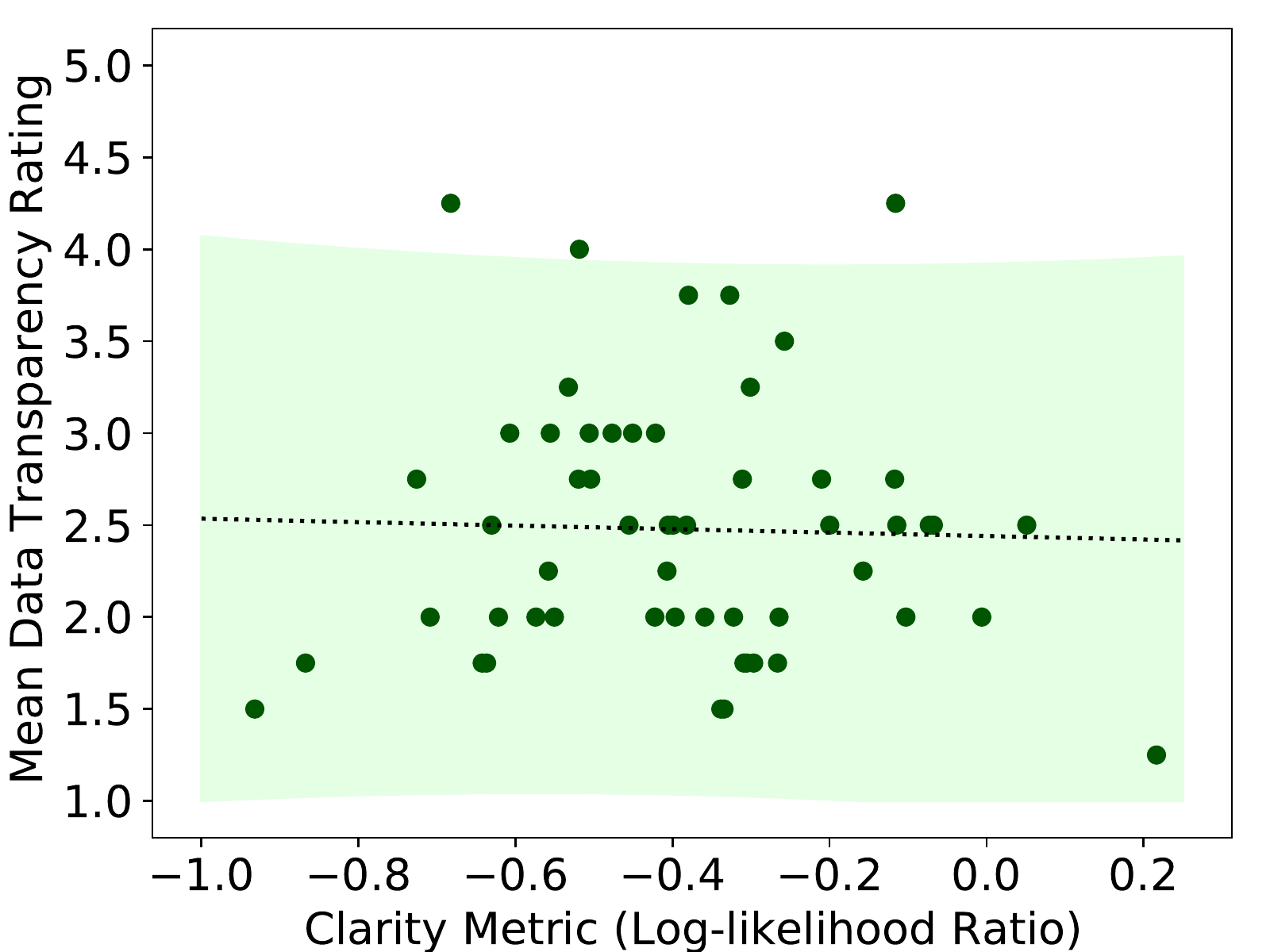}
\end{minipage}%
\begin{minipage}{.33\linewidth}
\centering
\includegraphics[width=1\textwidth]{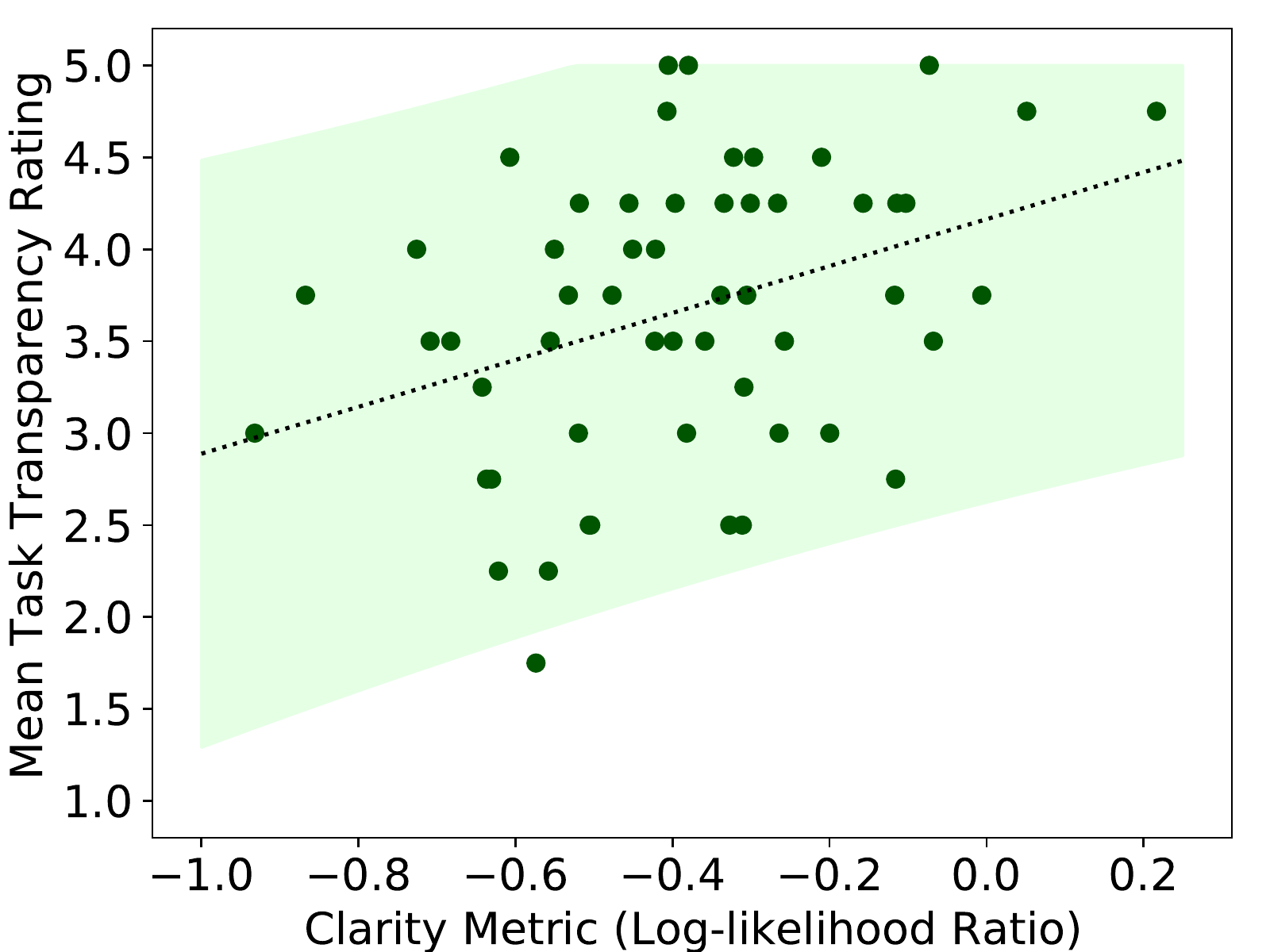}
\end{minipage}\\

\begin{minipage}{.33\linewidth}
\centering
\centering
\includegraphics[width=1\textwidth]{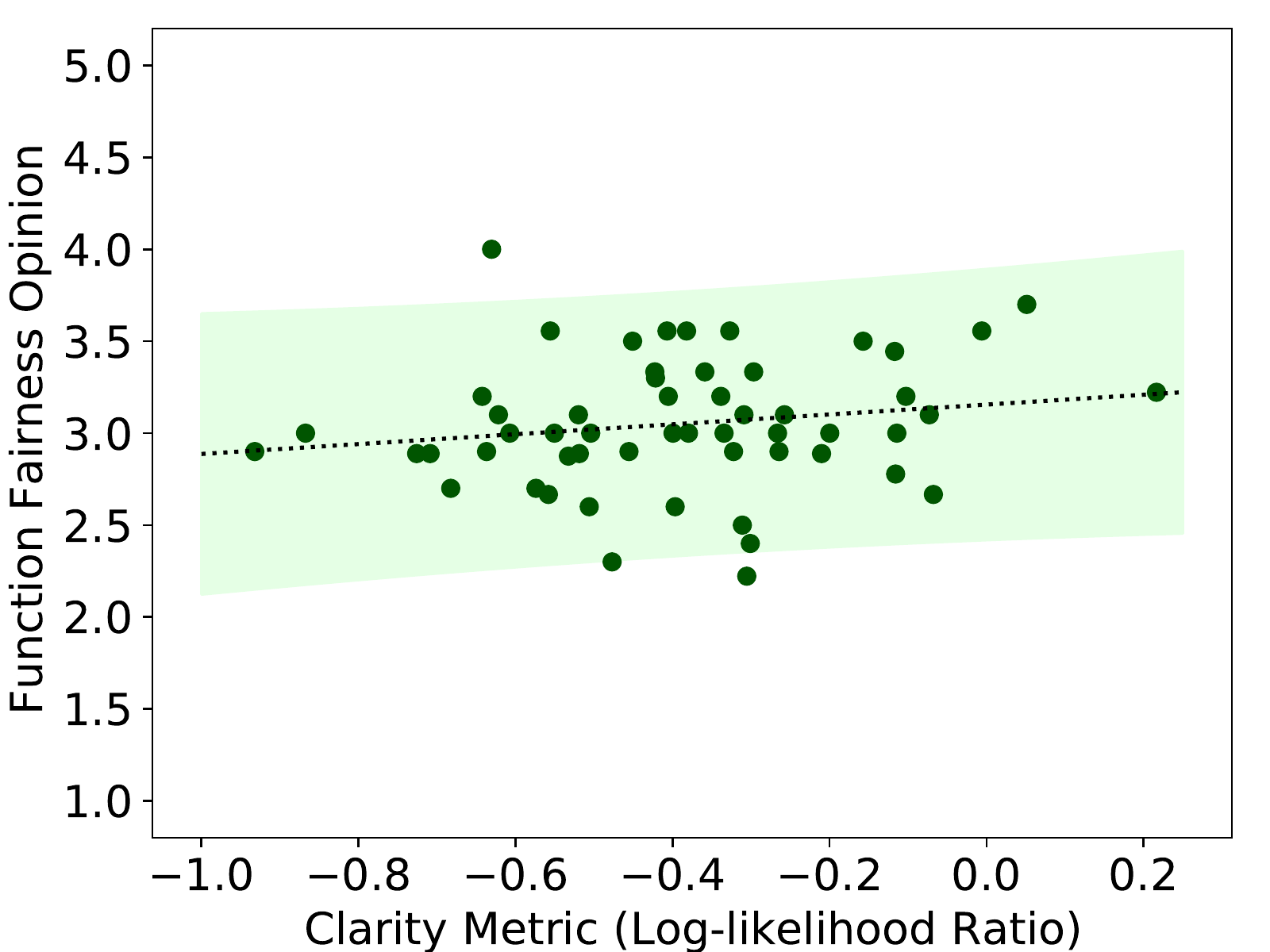}
\end{minipage}%
\begin{minipage}{.33\linewidth}
\centering
\includegraphics[width=1\textwidth]{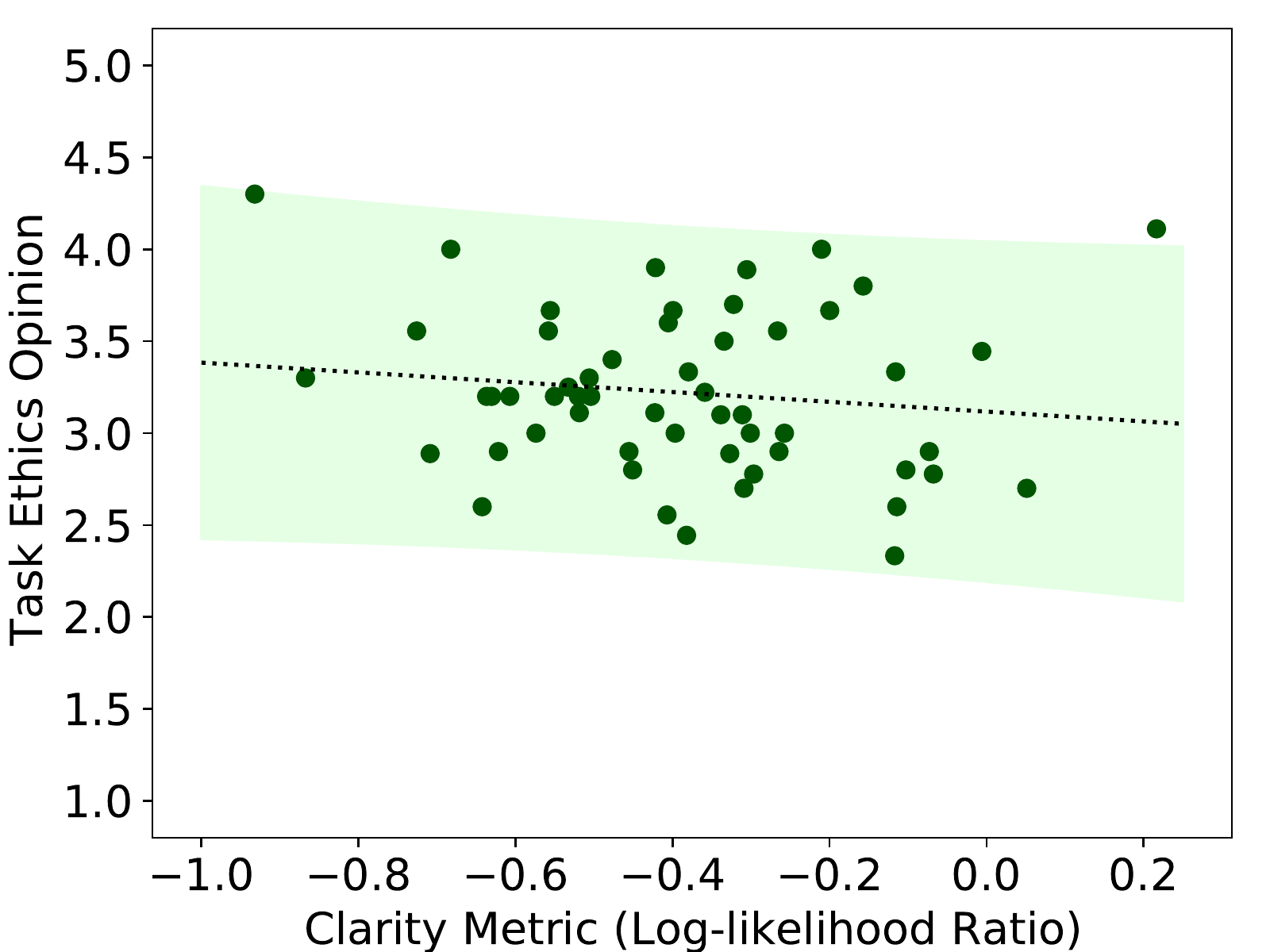}
\end{minipage}%
\begin{minipage}{.33\linewidth}
\centering
\includegraphics[width=1\textwidth]{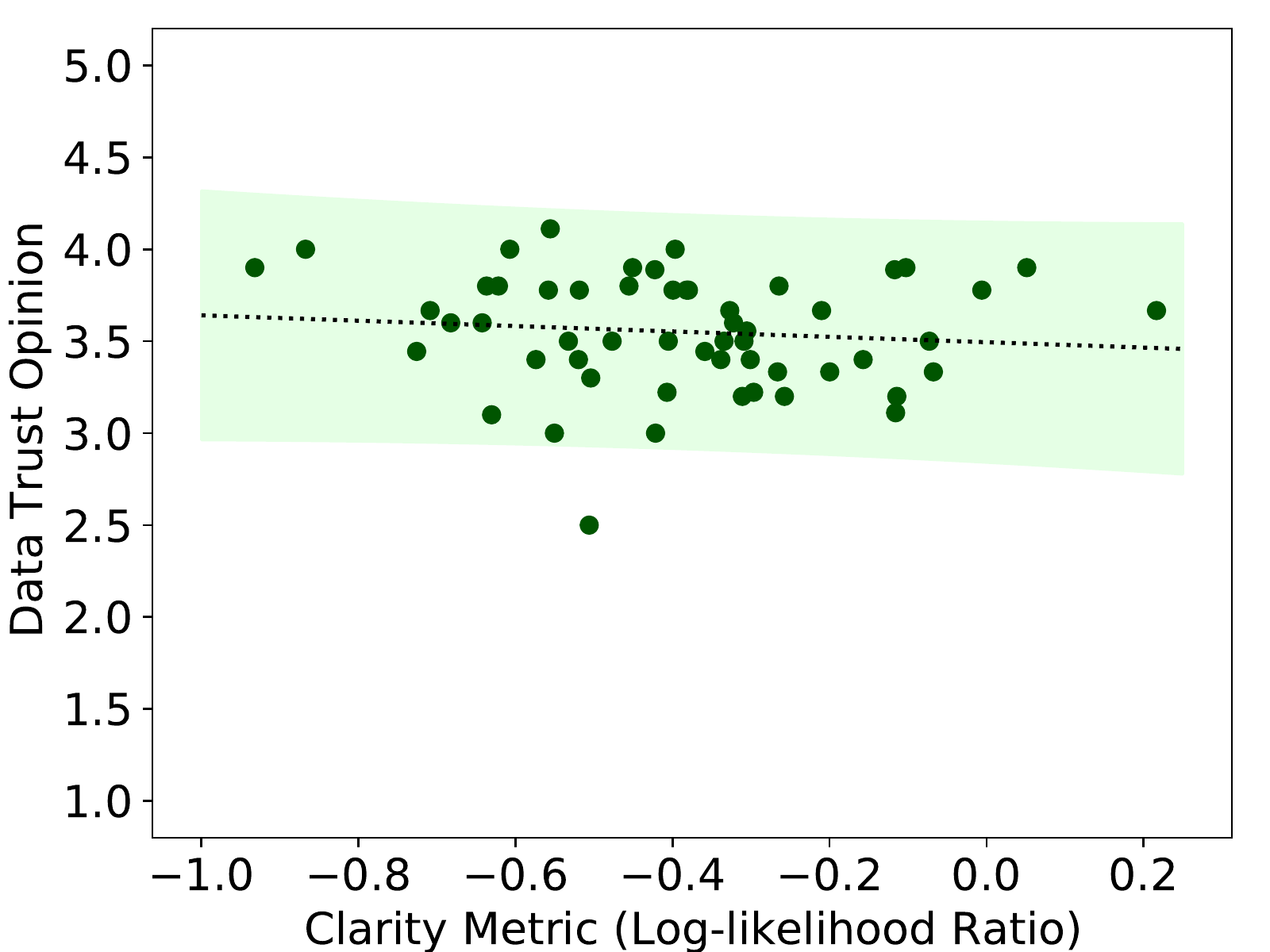}
\end{minipage}\\

\begin{minipage}{.33\linewidth}
\centering
\centering
\includegraphics[width=1\textwidth]{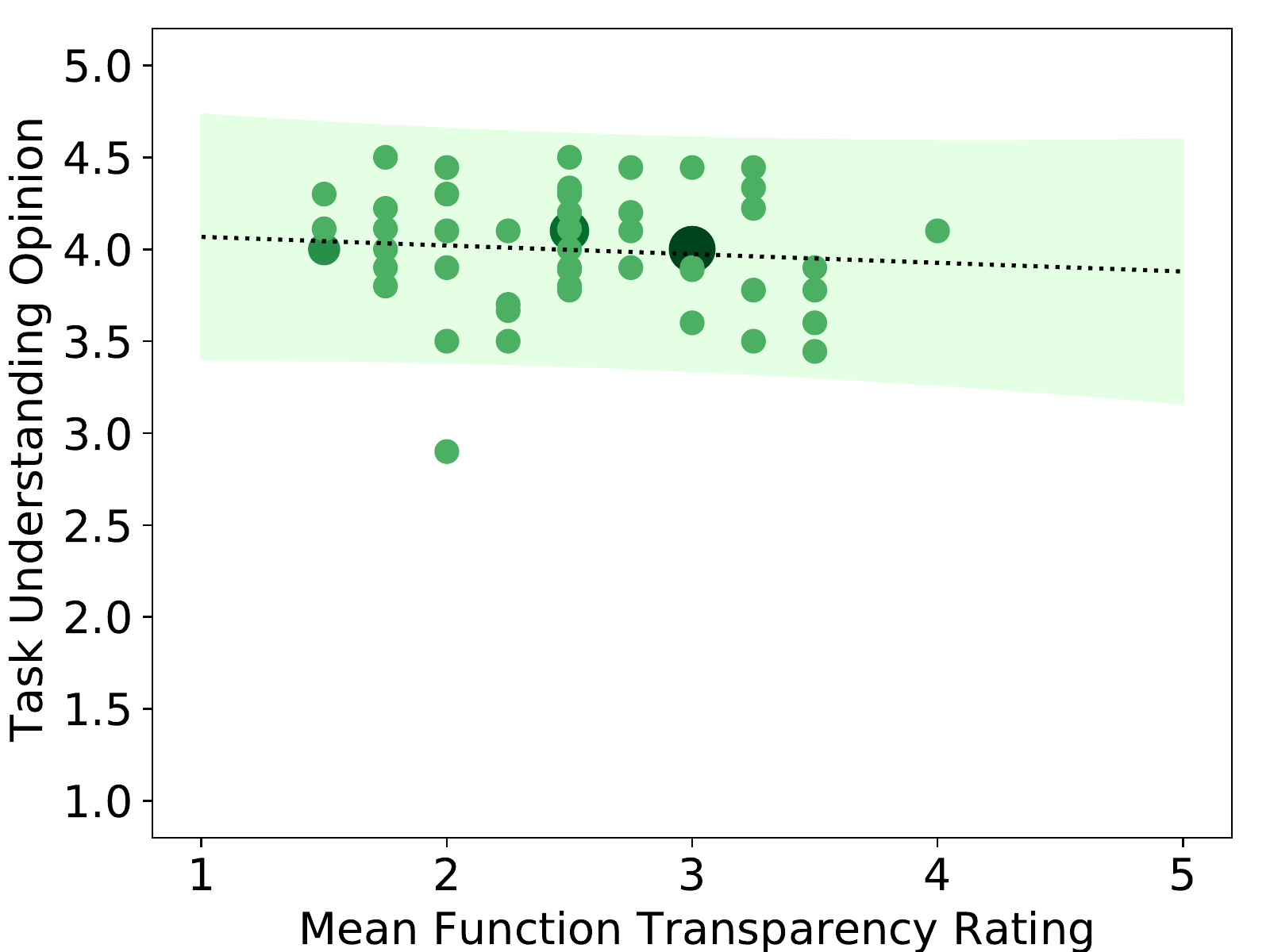}
\end{minipage}%
\begin{minipage}{.33\linewidth}
\centering
\includegraphics[width=1\textwidth]{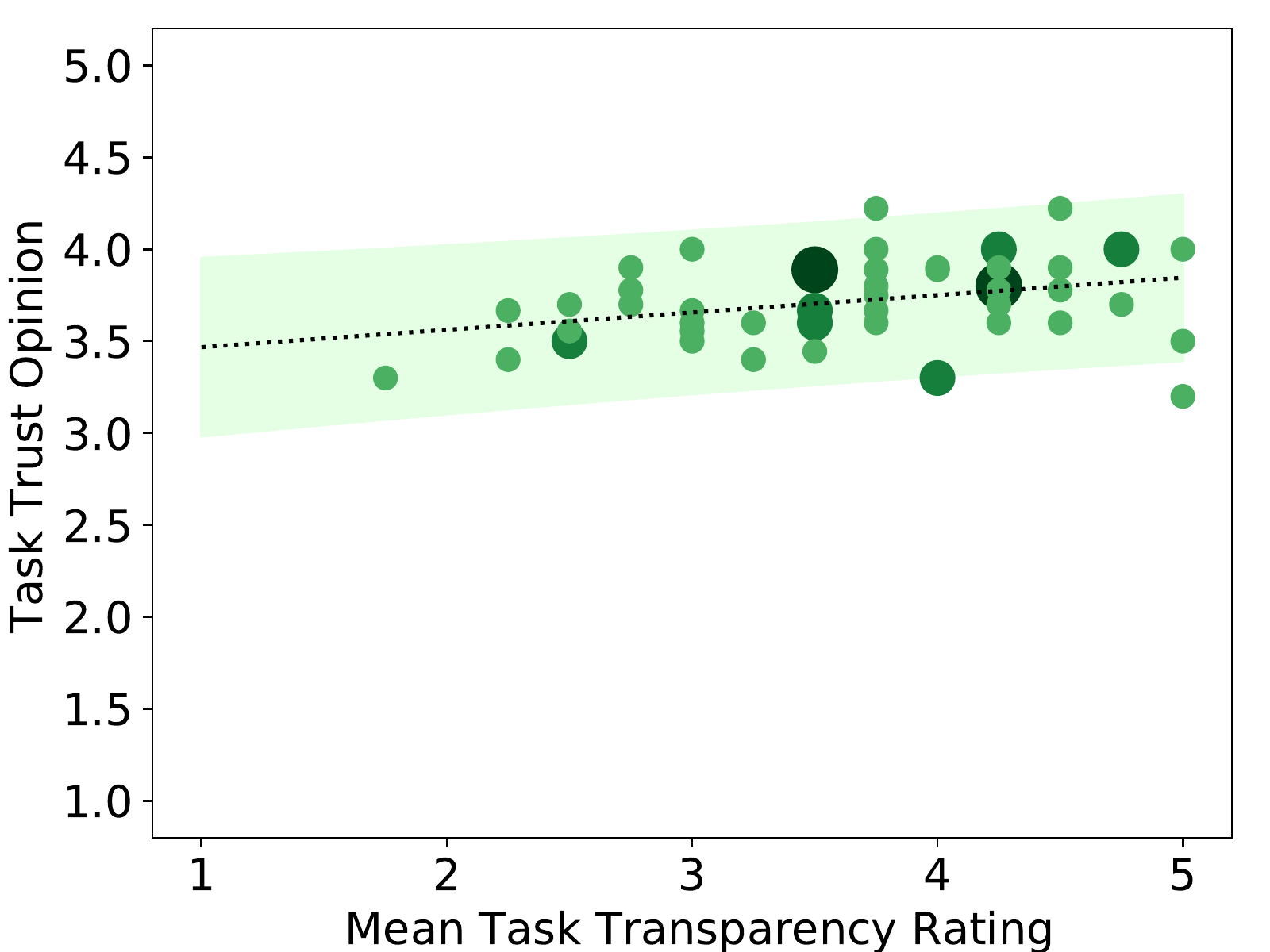}
\end{minipage}%
\begin{minipage}{.33\linewidth}
\centering
\includegraphics[width=1\textwidth]{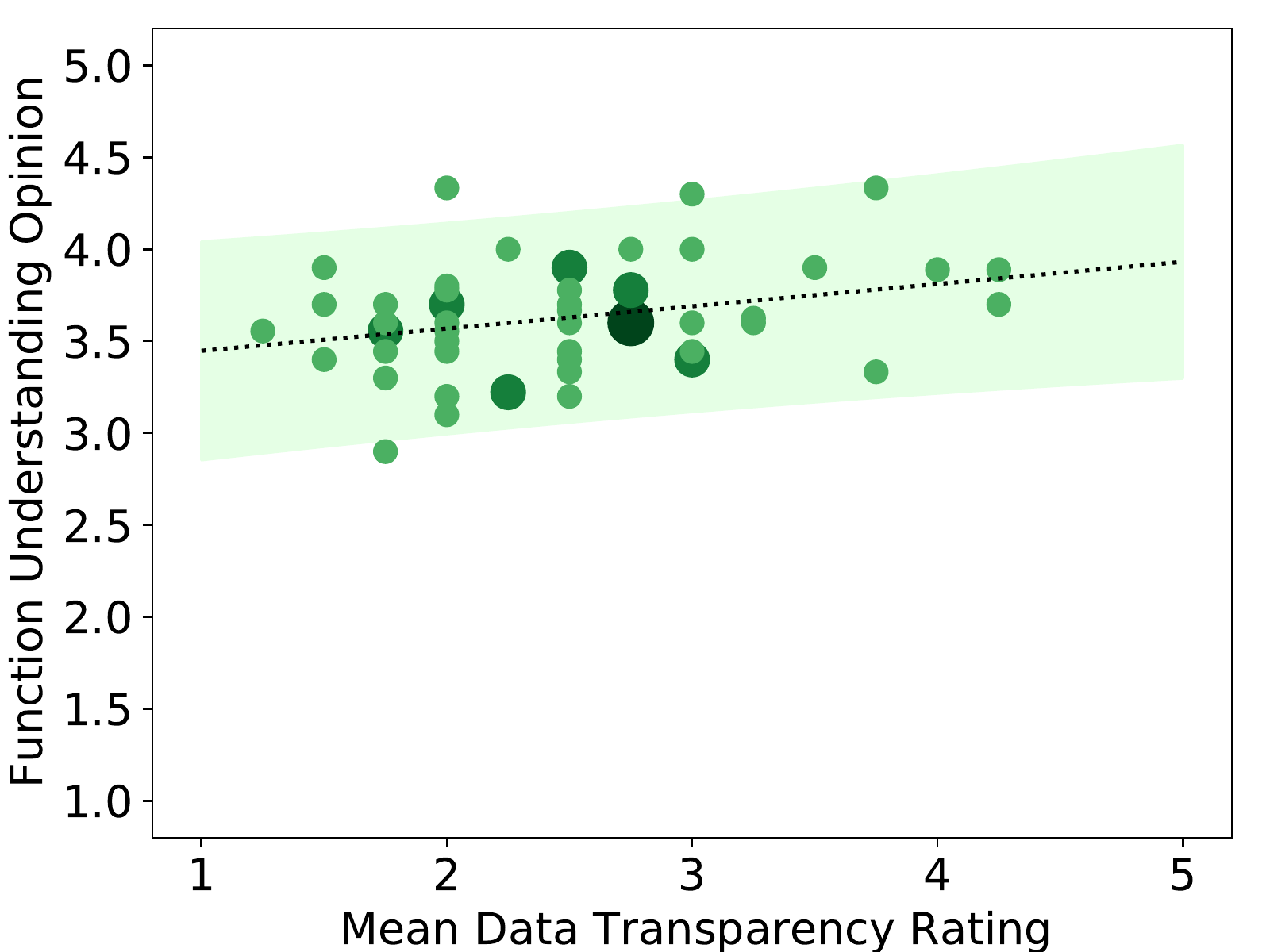}
\end{minipage}\\
	\caption{Scatterplots for other sets of bivariate analyses. Rows 1 and 2 compare the Replicability Metric to other response variables, rows 3 and 4 compare the S-A Metric to other response variables, and row 5 compares subjective expert transparency ratings to other response variables.}
	\label{fig:extra}
\end{figure*}

\end{document}